\documentclass[10pt,twocolumn,letterpaper]{article}

\usepackage{iccv}
\usepackage{times}
\usepackage{epsfig}
\usepackage{graphicx}
\usepackage{amsmath}
\usepackage{amssymb}

\usepackage{times}
\usepackage{epsfig}
\usepackage{graphicx}
\usepackage{amsmath}
\usepackage{amssymb}

\usepackage[utf8]{inputenc} % allow utf-8 input
\usepackage[T1]{fontenc}    % use 8-bit T1 fonts
\usepackage{url}            % simple URL typesetting
\usepackage{booktabs}       % professional-quality tables
\usepackage{enumitem}
\usepackage{amsfonts}       % blackboard math symbols
\usepackage{nicefrac}       % compact symbols for 1/2, etc.
\usepackage{microtype}      % microtypography
\usepackage{times}
\usepackage{epsfig}
\usepackage{graphicx}
\usepackage{comment}
\usepackage{multirow}
\usepackage{multicol}
\usepackage{pifont}
\usepackage{soul,color}
\usepackage{subcaption}

\usepackage{enumitem}
\usepackage[sort,numbers,square,comma]{natbib}
\usepackage[pagebackref=true,breaklinks=true,letterpaper=true,colorlinks,bookmarks=false]{hyperref}

\iccvfinalcopy % *** Uncomment this line for the final submission

\def\iccvPaperID{9227} % *** Enter the ICCV Paper ID here

\setitemize{noitemsep,topsep=0pt,parsep=0pt,partopsep=0pt}
\setlist{noitemsep,topsep=0pt,parsep=0pt,partopsep=0pt}
\renewcommand{\eg}{\emph{e.g., }}
\makeatletter
\renewcommand\paragraph{\@startsection{paragraph}{4}{\z@}%
  {.75ex \@plus1ex \@minus .2ex}% was 2.05ex then 0em
  {-1em}  % was -1em
  {\reset@font\normalsize\bfseries}
}
\newcommand{\glayout}{G^l}
\newcommand{\gstyle}{G^s}
\newcommand{\elayout}{E^l}
\newcommand{\estyle}{E^s}
\newcommand{\zlayout}{z^l}
\newcommand{\zstyle}{z^s}
\newcommand{\nmke }{NMKE}
\newcommand{\idsim }{ID-SIM}

\newcommand{\cc}[1]{}

\begin{document}

%%%%%%%%% TITLE
% \title{Two-step Synthesis: Few-shot Talking Heads via Learned Spatial Maps}
\title{Learned Spatial Representations for Few-shot Talking-Head Synthesis}
% \title{Supplementary Material for Learned Spatial Representations for Few-shot Talking-Head Synthesis}

% \and
\author{
  Moustafa Meshry\quad 
    Saksham Suri\quad 
    Larry S. Davis\quad
    Abhinav Shrivastava\\\\%[0.3em]
    University of Maryland, College Park\\ %\mbox{ }
}

\maketitle

%%%%%%%%% ABSTRACT
\begin{abstract}
    %% Abstract #1
    %% ------------
    % We propose a novel approach for few-shot talking-head synthesis.
    % Our method disentangles novel view synthesis into spatial and style components, and generates images in a two-step process.
    % The first step generates a spatial layout for the target image, and the second step utilizes spatial denormalization to achieve multi-modal image synthesis.
    % Disentangling spatial layout from style information allows the network to learn a strong prior of the shape distribution of a category of objects (such as human faces).
    % Furthermore, the predicted layout enables utilizing state-of-the-art image synthesis techniques, like spatial denormalization.
    % Our network learns a latent spatial representation that exhibits strong multi-view consistency and achieves state-of-the-art performance for novel view synthesis of talking heads.
    %% ------------------------------------------------------
    %% Abstract #1:
    %% ------------
    We propose a novel approach for few-shot talking-head synthesis.
    While recent works in neural talking heads have produced promising results, they can still produce images that do not preserve the identity of the subject in source images.
    We posit this is a result of the entangled representation of each subject in a single latent code that models 3D shape information, identity cues, colors, lighting and even background details.
    In contrast, we propose to factorize the representation of a subject into its spatial and style components.
    Our method generates a target frame in two steps. First, it predicts a dense spatial layout for the target image. Second, an image generator utilizes the predicted layout for spatial denormalization and synthesizes the target frame.
    We experimentally show that this disentangled representation leads to a significant improvement over previous methods, both quantitatively and qualitatively.
\end{abstract}

% %%%%%%%%% BODY TEXT
\section{Introduction}
\label{sec:intro}

%% Problem statement:
%% ------------------
We study the task of learning personalized head avatars in a low-shot setting, also known as \emph{``neural talking heads''}.
Given a single-shot or few-shot images of a source subject, and a driving sequence of facial landmarks, possibly derived from a different subject, the goal is to synthesize a photo-realistic video of the source subject, under the poses and expressions of the driving sequence.
%
% Given a single-shot or few-shot images of a source subject, and a driving sequence of facial landmarks, possibly derived from a different subject, the
%
% We consider landmark-driven talking-head synthesis.
% The inputs are a single-shot or few-shot images of a target subject, as well as a driving sequence of facial landmarks, possibly derived from a different subject.
% And the goal is to synthesize a realistic head avatar under the poses and expressions of the driving sequence.
%
%% Motivating the problem/task:
%% ----------------------------
This task has a wide range of applications, including those in AR/VR, video conferencing, gaming, animated movie production and video compression in tele-communication.

%% Brief history of previous approaches:
%% -------------------------------------
Traditional graphics-based approaches to this task rely on a 3D face geometry and produce very high quality synthesis. 
However, they tend to focus on modeling the face area without the hair, and they learn a subject-specific model and cannot generalize to new subjects.
In contrast, recent 2D-based approaches~\cite{zakharov2019few,Siarohin_2019_NeurIPS,burkov2020neural,Zakharov20fast} learn a subject-agnostic model that can animate unseen subjects given as few as a single image.
Furthermore, since these works learn an implicit model and do not require an explicit geometric representation, they can synthesize the full head, including the hair, mouth interior, and even wearable accessories like glasses and earrings.
This remarkable generalization ability however comes at the cost of low quality and poor identity preservation when compared to their 3D-based subject-specific counterparts.
Bridging the quality gap between 2D-based subject-agnostic and 3D-based subject-specific approaches remains an open problem.

% One class of 2D approaches 
Recent efforts in 2D-based approaches can be divided into two classes; \emph{warping-based} and \emph{direct synthesis}.
%%% 1. warping-based approaches:
As the name suggests, warping-based methods (\eg\cite{Siarohin_2019_NeurIPS}) learn to warp the input image or a recovered canonical pose based on the motion of the driving sequence.
While these methods achieve high realism, especially for static and rigid parts of the image, they tend to work well only for a limited range of motion, head rotation and dis-occlusion.
%%% 2. direct synthesis approaches:
% -- Direct synthesis approaches on the other hand can handle a wide range of poses and expressions.
% -- This is due to learning a strong prior of human faces in a highly compressed latent space.
% -- Direct synthesis utilize i2i networks and in turn are impacted by advances in the i2i literature.
%
%<<<<<<<<<<<<<<<<<<<< TEASER
\begin{figure}[t!]
    \centering
    \includegraphics[width=0.99\linewidth]{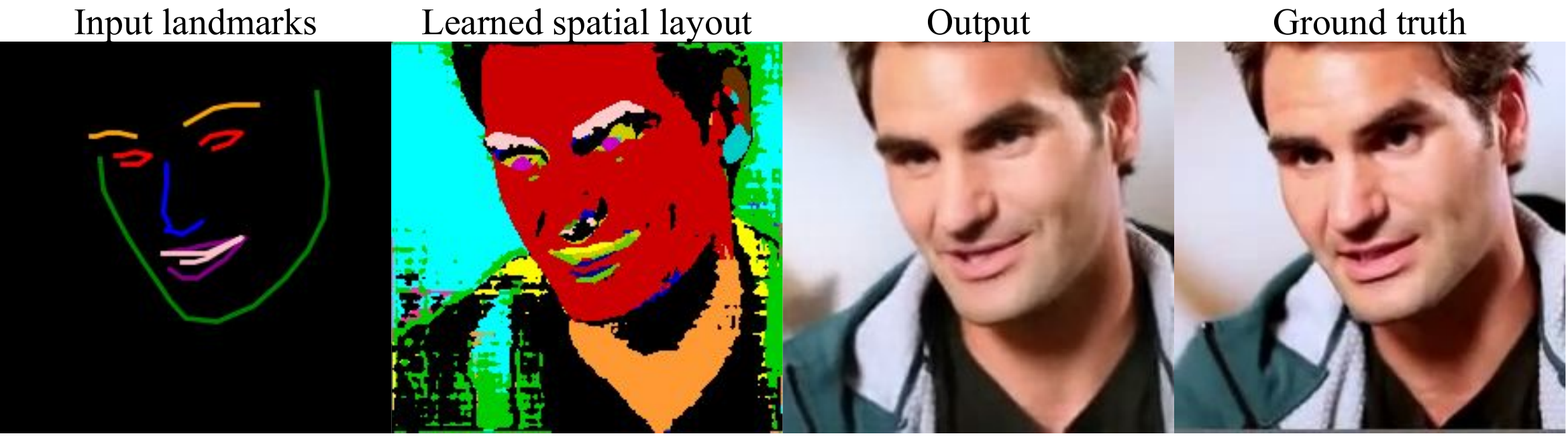}
    % \vspace{-0.4cm}
    \caption{Our framework disentangles spatial and style information for image synthesis. It predicts a latent spatial layout for the target image, which is used to produce per-pixel style modulation parameters for the final synthesis.}
    % \vspace{-0.4cm}
    \label{fig:teaser}
\end{figure}

%>>>>>>>>>>>>>>>>>>>>>>>>>>>>
%<<<<<<<<<<<<<<<<<<< PIPELINE
\begin{figure*}[t!]
    \centering
    \includegraphics[width=0.85\linewidth]{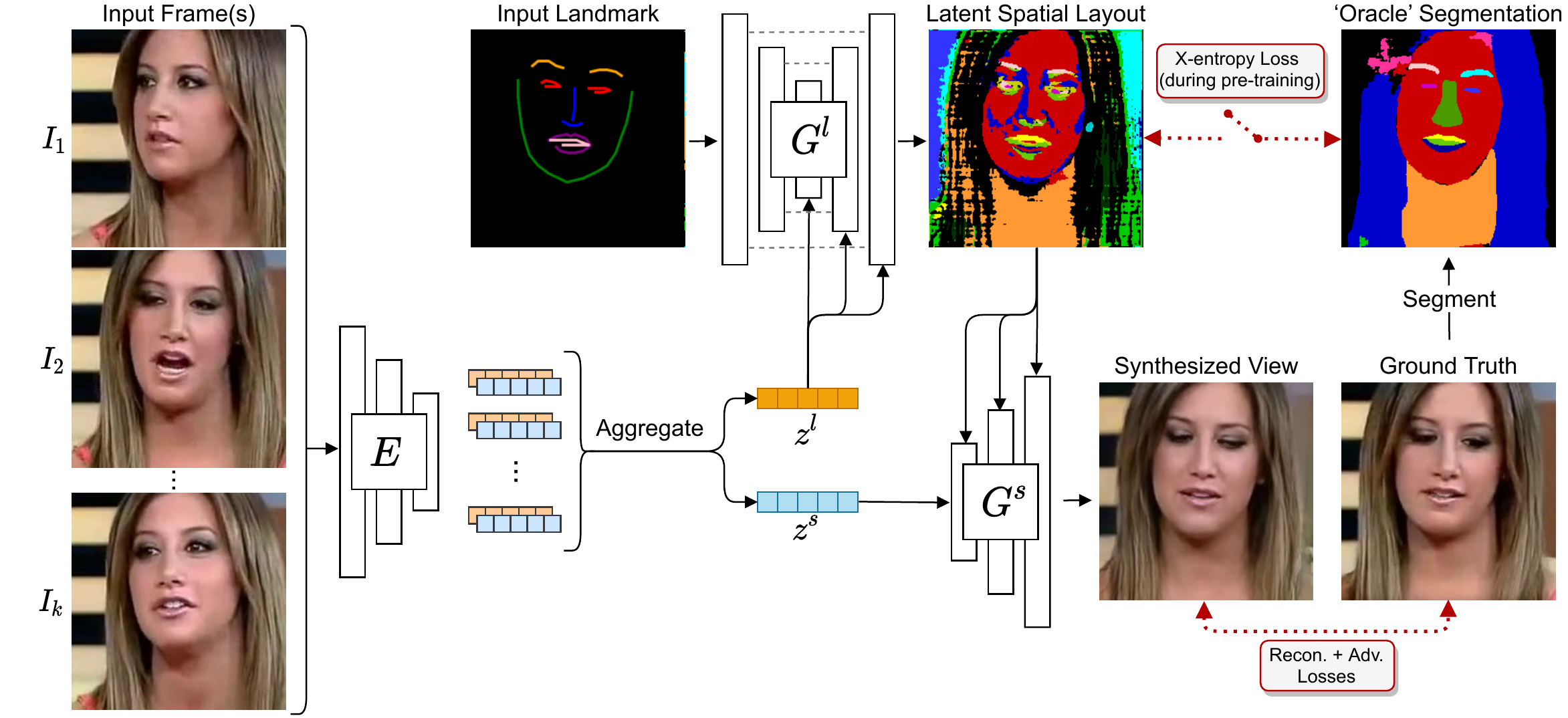}
    \caption{Overview of our training pipeline. The cross-entropy loss with the oracle segmentation is used during pre-training the layout predictor $\glayout$, and then turned off during the full pipeline training.}
    \label{fig:training}
\end{figure*}

%>>>>>>>>>>>>>>>>>>>>>>>>>>>>
%
On the other hand, direct synthesis approaches (\eg\cite{zakharov2019few,burkov2020neural,Zakharov20fast}) 
encode the source subject into a compressed latent code, and a generator decodes the latent code to synthesize the target pose.
These approaches learn a prior over the compressed latent space, and can generate realistic results for a wider range of poses and head motion.
However, they exhibit a noticeable identity gap between their output and the source subject.
%, and rely on a subject-specific fine-tuning to improve their result.

We posit that the identity gap is caused by the entangled representation of the source subject in a single latent code.
This compressed 1D latent encodes multi-view shape information, identity cues, as well as color information, lighting and background details.
In order to synthesize a target view from a latent code, the generator needs to devise a complex function to decode the uni-dimensional latent into its corresponding 2D spatial information.
We argue this not only consumes a large portion of the network capacity, but also limits the amount of information that can be encoded in the latent code.
% On the other hand, conditioning the generator on an accurate and dense spatial map (such as a semantic labeling) alleviates much of this burden from the generator and results in an improved performance.
% Furthermore, recent advances in image translation architectures~[??,??,??] utilize input semantic maps to generate per-pixel denormalization parameters, which resulted in superior performance in I2I translation compared to traditional UNet and encoder-decoder architectures.
% While semantic segmentation maps for novel views of the target subject do not exist in our task, our predicted layout maps act as a spatial proxy that enables leveraging spatial denormalization architectures to our favor.

To address this problem, we propose a two-step framework that decomposes the synthesis of a talking head into its spatial and style components.
Our framework animates a source subject in two steps.
First, it predicts a novel spatial layout of the subject under the target pose and expression.
Then, it synthesizes the target frame conditioned on the predicted layout.
This factorized representation yields the following key performance advantages.
%
% \\
% \noindent
% \textbf{Better subject-agnostic model performance.}
\paragraph{Better subject-agnostic model performance.}
The performance of our subject-agnostic (also called \emph{meta-learned}) model not only performs better than previous subject-agnostic state-of-the-art, but is also on-par with the subject-finetuned performance of previous works when there are only few source images available (\eg less than 10 images).
%
% \\
% \noindent
% \textbf{Better fine-tuned performance with less data.}
\paragraph{Better fine-tuned performance with less data.}
Fine-tuning our model for a specific subject requires significantly less data and fewer iterations than previous works, and yet achieves better performance. For example, we show that fine-tuning our model using 4-shot inputs outperforms previous state-of-the-art models fine-tuned using 32-shot inputs.
%
% \\
% \noindent
% \textbf{Robustness to pose variations.}
\paragraph{Robustness to pose variations.}
We show that our model is more robust against a wider range of poses and facial expressions, while still producing both realistic and identity-preserving results.
%
% \\
% \noindent
% \textbf{Improved identity preservation.}
\paragraph{Improved identity preservation.}
Shape difference between the source and driving identities poses a challenge for identity preservation in reenacted results. The intermediate novel spatial representation learned by our model reduces the sensitivity towards such differences and better preserves the identity.
%
% \begin{itemize}[leftmargin=*]
%     \item Significant performance improvement over previous few-shot methods along two axes:
%     \begin{itemize}[leftmargin=*]
%         \item First, our method bridges the gap between subject-agnostic and subject-finetuned models. Specifically, we show that the meta-learning performance of our model is on-par with the subject fine-tuned performance of previous works when only few source images are available (\eg less than 10 images).
%         \item Second, fine-tuning our model for a specific subject requires significantly less data and fewer iterations and yet achieves better performance. For example, we show that fine-tuning our model using 4-shot inputs outperforms previous state-of-the-art models fine-tuned using 32-shot inputs.
%     \end{itemize}
%     \item Our model is more robust against a wider range of poses and facial expressions, while still producing both realistic and identity-preserving results.
%     \item The intermediate novel spatial representation reduces the sensitivity towards the face shape difference between the source and driving identities, which poses a challenge for identity preservation for reenacted results.
% \end{itemize}
%

\noindent
In summary, we make the following contributions:
\begin{itemize}[leftmargin=*]
    \item A novel approach that disentangles the spatial and style components for \emph{talking-head} synthesis.
    \item A novel latent spatial representation that proves effective for few-shot novel view synthesis.
    \item We achieve state-of-the-art performance in both the single-shot and multi-shot settings, as well as in the meta-learned and subject-finetuned modes.
\end{itemize}

\section{Related work}
\label{sec:related_work}
Existing approaches for realistic talking-head synthesis can be categorized into 3D-based and 2D-based.
\paragraph{3D-based methods.} 
Such methods~\cite{thies2015real,thies2016face2face,suwajanakorn2017synthesizing} utilize 3D geometric representations as a proxy to animate a target subject.
Common geometric representations, such as 3D morphable models (3DMM)~\cite{blanz1999morphable}, only model the face area, and do not include challenging regions like the hair, eyes and mouth interior. Obtaining a detailed geometry of these regions is an expensive and challenging task.
Therefore, such methods either cannot synthesize or perform poorly on those regions.
Recent works~\cite{kim2018deepvideoportraits,thies2019deferred,gafni2020dynamic} combine the traditional graphics pipeline with machine learning to better model the eye movement, mouth interior, or learn a better appearance model.
However, they learn subject-specific models that do not generalize to new subjects.
Other works~\cite{nagano2018pagan,fried2019text} take first steps to generalize to multiple subjects but they do not perform well on hair and other regions outside the face.
\paragraph{2D-based methods.} These methods~\cite{chen2019hierarchical,nirkin2019fsgan,pumarola2018ganimation,gu2020flnet,ha2020marionette,Zakharov20fast,burkov2020neural,Siarohin_2019_NeurIPS,wang2018fewshotvid2vid,zakharov2019few} learn an implicit model of the head and do not require a proxy geometry.
Therefore they can synthesize the full head including dynamic regions like the hair, eyes, and mouth interior. They can also model different wearable accessories such as hats, glasses, and earrings.
Early works build on top of CycleGAN~\cite{zhu2017unpaired} and learn subject-specific models~\cite{bansal2018recycle,wu2018reenactgan}.
More recent works~\cite{zakharov2019few,wang2018fewshotvid2vid,Siarohin_2019_NeurIPS,burkov2020neural,ha2020marionette,Zakharov20fast} learn subject agnostic models that can animate unseen subjects given only a single or few-shot images.
However, these methods lack in quality and identity preservation compared to the 3D-based subject-specific models.
To bridge this performance gap, hybrid models~\cite{zakharov2019few,burkov2020neural,Zakharov20fast} utilize a meta-learning phase that trains a subject-agnostic model on a large corpus of data, then an optional subject-specific fine-tuning phase is performed to improve the realism and restore the source identity.
%
% In this work, we improve the meta-learned performance to achieve state-of-the-ar
% Our proposed method falls into this category. We improve the performance of the meta-learned model, while also benefiting from the optional fine-tuning phase to further refine our output.
% In this work, we significantly improve the meta-learned performance compared to previous work.
In this work, we improve the meta-learned performance to achieve state-of-the-art results without any subject-specific fine-tuning.
While our model could still benefit from the optional fine-tuning phase to further refine the results, it requires significantly less data samples compared to previous works.
% We also show that our model could still benefit from the optional fine-tuning phase to refine our results, while requiring significantly less data samples compared to previous works.
%

On another axis, 2D-based approaches can be categorized based on the synthesis technique into warping-based (\eg\cite{wiles2018x2face,Siarohin_2019_NeurIPS,ha2020marionette,wang2020one}) and direct synthesis (\eg\cite{zakharov2019few,burkov2020neural,Zakharov20fast}).
Warping-based approaches warp an input image~\cite{Siarohin_2019_NeurIPS,ha2020marionette} or a recovered canonical pose~\cite{wiles2018x2face} to synthesize novel poses.
Warping results however tend to break when the target pose is far from that of the source image.
Direct synthesis approaches utilize advances in Generative Adversarial Networks (GANs)~\cite{goodfellow2014generative} and Image-to-Image (I2I) translation~\cite{isola2017image} to generate novel poses.
Compared to warping-based approaches, direct synthesis methods can realistically handle a wider range of poses and expressions.
%, although they require a subject-specific fine-tuning to preserve the identity.

\paragraph{Multi-modal Image-to-Image (I2I) translation.}
Several multi-modal I2I translation works feed a style latent code, either directly to the generator~\cite{zhu2017toward} or through adaptive instance normalization (AdaIN) ~\cite{huang2017arbitrary,huang2018multimodal}.
Recent state-of-the-art architectures~\cite{park2019SPADE,liu2019learning,zhu2020sean} showed a significant improvement over traditional UNet~\cite{ronneberger2015u} and encoder-decoder architectures, by generating per-pixel spatial denormalization (SPADE) parameters~\cite{park2019SPADE}.
However, such architectures depend on the existence of accurate semantic segmentations or other dense spatial representations of the target image, hence limiting their usage in tasks where such dense representations do not exist.
In this work, we learn to predict a latent dense layout to provide the spatial input to SPADE.

\section{Method}
\label{sec:approach}

%--------------------------------
% \input{figures_tex/architecture}
% \input{figures_tex/training}
\begin{figure}[t!]
    \centering
    \includegraphics[width=0.99\linewidth]{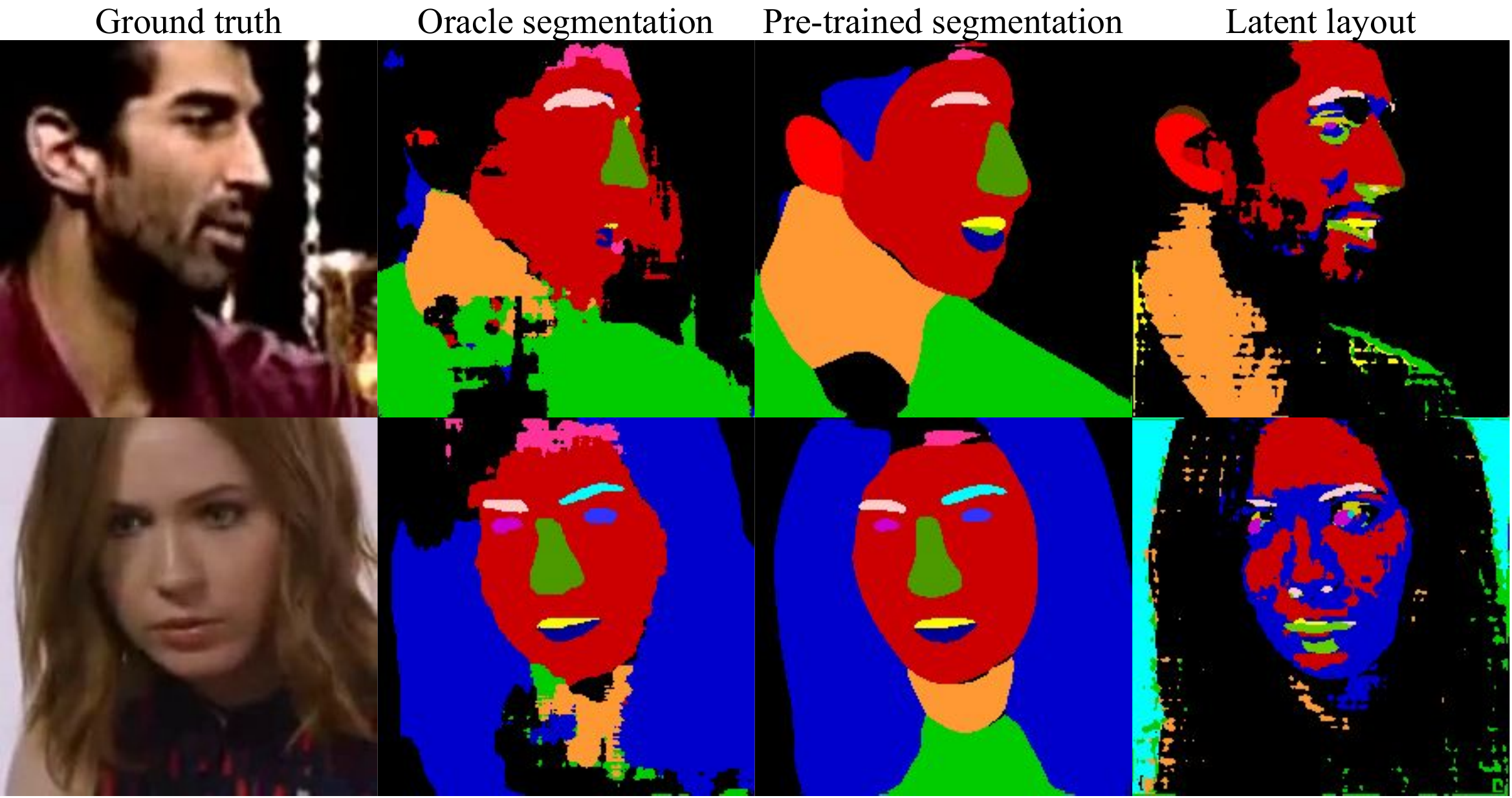}
    % \vspace{-0.4cm}
    \caption{Layout pre-training predicts meaningful segmentation maps despite the noisy oracle segmentations. Our latent spatial representation encodes more information than traditional segmentations.}
    % \vspace{-0.5cm}
    \label{fig:layout}
\end{figure}

%--------------------------------

Our approach factorizes the representation of a head avatar into spatial and style components.
It breaks down the novel view head synthesis of a subject into two steps.
% a spatial and style components.
First, a layout prediction network $\glayout$ translates facial landmarks for a target view into a dense spatial layout of the subject.
Then, an image generator $\gstyle$ synthesizes the final image conditioned on the predicted layout.
% First, we predict a dense spatial layout for the target view, and then we synthesize the final image.
% For simplicity of illustration, we’ll consider semantic maps as the dense spatial representation.
%
% In the following, 
We first give an overview of our pipeline in Section~\ref{sec:overview}.
Then, we explain how to pre-train the layout prediction network $\glayout$ to predict semantic segmentations of novel views in Section~\ref{sec:layout_prediction}, followed by the full pipeline training in Section~\ref{sec:training}.
Section~\ref{sec:latent_layout} explains how the layout prediction network $\glayout$ transitions from predicting semantic maps to learning a more powerful latent spatial representation.
And finally, we discuss how to learn a personalized head avatar through an optional subject-specific fine-tuning stage in Section~\ref{sec:finetuning}.

% ------------------------------------------------------------------------------------
\subsection{Overview}
\label{sec:overview}
Given $K$-shot inputs $\{I_1 \dots I_K\}$ of a source subject, a two-headed encoder $E = \{\elayout, \estyle\}$ processes the inputs and generates $K$ layout latents $\{\zlayout_{i}\}$ and $K$ style latents $\{\zstyle_{i}\}$ for $i \in \{1 \dots K\}$.
The $K$ latents are then averaged to get an aggregated layout latent $\zlayout = \frac{1}{K}\sum_{i=1}^K \zlayout_i$ and style latent $\zstyle = \frac{1}{K}\sum_{i=1}^K \zstyle_i$.
Averaging the $K$ latents cancels out view-specific information and transient occluders, and maintains implicit 3D information like the head and hair shape for the layout latent, and color and lighting information for the style latent.
We have two generators: a layout predictor network $\glayout$ and an image generator $\gstyle$.
The layout predictor takes as input the facial landmarks for a target view $x_t$ and the layout latent $z^l$ and generates a spatial layout $y^l_t = \glayout(x_t, z^l)$, such as a semantic map, for the target view.
% The image generator $\gstyle$ is a SPADE generator~\cite{park2019SPADE} that takes as input the style latent $\zstyle$ and is spatially conditioned on the predicted layout $y^l$ and synthesizes the final image as $\hat{I} = \gstyle(y^l, z^s)$.
The image generator $\gstyle$ processes the style latent $\zstyle$ and utilizes spatial denormalization layers (SPADE~\cite{park2019SPADE}), conditioned on the predicted layout $y^l$, to synthesize the final image $\hat{I} = \gstyle(y^l, z^s)$.
An overview of our framework is shown in Figure~\ref{fig:training}.

% ------------------------------------------------------------------------------------
\subsection{Layout prediction pre-training}
\label{sec:layout_prediction}
Training the above pipeline end-to-end without any supervision or constraints on the predicted layouts results in a degenerate solution, where the spatial layouts and their corresponding spatial denormalization are completely ignored.
All spatial and style information are thus encoded into and decoded from the style latent $\zstyle$, which results in a poor performance.
Therefore, we opted to pre-train the layout prediction network to predict a plausible semantic segmentation of a target view, given the input facial landmarks $x_t$ and the layout latent $\zlayout$. % encoded from the few-shot inputs $\{I_i, \dots I_K\}$.
To supervise this training, we use an off-the-shelf face segmentation network~\cite{CelebAMask-HQ} as an oracle to segment the target image $I_t$ into a semantic map $S_t$, and we apply a cross-entropy loss (X-ent) between the oracle segmentation $S_t$ and our predicted segmentation $y^l_t = \glayout(x_t, z^l)$.
%
% We start by pre-training a semantic layout prediction network $\glayout$. 
% We generate pseudo ground truth segmentation obtained by segmenting training images using an off-the-shelf segmentation network~\cite{CelebAMask-HQ}, and train $G^l$ using a cross-entropy loss.
%
We observe that the obtained oracle segmentations are very noisy and have poor quality (\eg Figure~\ref{fig:layout}).
This is caused by the domain gap, in terms of image resolution and the distribution of head poses, between the datasets used to train the oracle segmentation network~\cite{CelebAMask-HQ}, and in-the-wild videos of talking heads.
Thus, to regularize the segmentation prediction training, we use a mutli-task pre-training strategy where the layout prediction network predicts an extra RGB reconstruction $R_t$ of the target image $I_t$, which is used as a secondary supervisory signal. Specifically, we have
\begin{equation}
    y^l_t, R_t = \glayout(x_t, \zlayout), \qquad \zlayout = \frac{1}{K}\sum_{i=1}^K E^l(I_i)
\end{equation}
And the objective for the pre-training is
\begin{equation}
    \mathcal{L}_{\text{seg}} = \text{X-ent}(y^l_t, S_t) + \lambda_R \mathcal{L}_R(R_t, I_t)
\end{equation}
where $\mathcal{L}_R$ is a perceptual reconstruction loss, and $\lambda_R$ is a relative weighting term which is set to a low value.
%
% After pre-training $G^l$, we plug it in and train the full pipeline which generates a target view as $\hat{I} = G^s(G^l(x_t, z^l), z^s)$.
% %
% We explore three variants:
% \begin{itemize}
%     \item {\emph{fixed seg.:}} we keep the weights of the pre-trained layout  prediction network $G^l$ fixed.
%     \item {\emph{ft. seg.:}} we allow the weights of $G^l$ to be fine-tuned with the full pipeline, while still supervising the predicted segmentations with a      cross-entropy loss.
%     \item {\emph{latent spatial maps:}} we fine-tune $G^l$ with the full  pipeline, but without supervising the predicted segmentations. This allows the predicted layout to deviate from a semantic representation to other latent representations that better suits the image synthesis task at hand.
% \end{itemize}

% ------------------------------------------------------------------------------------
\subsection{Full pipeline training}
\label{sec:training}
Once the layout predictor network has been pre-trained to predict semantic segmentations, we plug it into our full pipeline. The predicted segmentation is fed as the spatial input to a SPADE image generator $\gstyle$ that synthesizes the final image as
\begin{equation}
    \hat{I} = G^s(G^l(x_t, z^l), z^s), \qquad \zstyle = \frac{1}{K}\sum_{i=1}^K E^s(I_i)
\end{equation}
We observe that the SPADE generator quickly utilizes the input spatial segmentations to resolve spatial ambiguities, and we no longer fall into a degenerate solution where the spatial input is ignored.

Our full pipeline, comprising the layout and style encoders $\{E^l, E^s\}$, the layout predictor $\glayout$ and the image generator $\gstyle$, is optimized to minimize three losses; a reconstruction loss $\mathcal{L}_{\text{rec}}$, an adversarial loss $\mathcal{L}_{\text{adv}}$, and a latent regularization loss $\mathcal{L}_{L2}$.

For the reconstruction loss $\mathcal{L}_{\text{rec}}$, we employ a perceptual loss~\cite{johnson2016perceptual} based on both the VGG19~\cite{simonyan2014very} and VGGFace~\cite{parkhi2015deep} networks, as well as an $L1$ loss. 
While the VGG19-based perceptual loss is a standard reconstruction loss, we follow Zakharov \etal~\cite{zakharov2019few} and utilize a VGGFace-based perceptual loss to promote identity preservation. We also use an $L1$ loss to better preserve color transfer between the synthesized and ground truth images.

The adversarial loss, $\mathcal{L}_{\text{adv}}$, encourages the output to be photo-realistic. To achieve that, a discriminator network $D$ is trained to discriminate between real and fake images, while the generator network, $\gstyle$ aims to fool the discriminator by bringing the output closer to the manifold of real images.
We borrow the architecture of the discriminator network $D$ from~\cite{karras2020analyzing} and use a non-saturating logistic loss with gradient penalty~\cite{mescheder2018training}.
Finally, we impose an $L2$ regularization on the learned latent codes to encourage compactness of the latent space.
The full training objective is given by

\begin{equation}
    \label{eqn:objective}
    \begin{split}
        \min{} & \mathcal{L} (\hat{I}_t, I_t, \zlayout, \zstyle| E^l, E^s, G^l, G^s, D) = \mathcal{L}_{\text{rec}}(\hat{I}_t, I_t) + \\
         & \quad \lambda_{\text{adv}}\mathcal{L}_{\text{adv}}(\hat{I}_t, I_t) + \lambda_{L2}\big(\|\zlayout\|^2 + \|\zstyle\|^2\big)
    \end{split}
\end{equation}
where $\lambda_{\text{rec}}, \lambda_{L2}$ determine the relative weights between the loss terms.

% ------------------------------------------------------------------------------------
\subsection{Learning a latent spatial representation}
\label{sec:latent_layout}
Spatial denormalization (SPADE) generates per-pixel denormalization parameters by feeding a dense spatial input through a small convolutional subnetwork.
While SPADE~\cite{park2019SPADE} originally uses semantic maps as input, we explore learning a latent spatial representation that better suits the image synthesis task at hand.
To do this, we turn off the cross-entropy loss so as to give the layout predictor $\glayout$ the freedom to diverge from predicting traditional semantic segmentations and learn other latent representations that better optimize the few-shot novel view synthesis objective.
The layout predictor is thus supervised only by the training objective of Eqn.~\ref{eqn:objective}.
Figure~\ref{fig:layout} shows examples of the learned latent layouts.
Although they might look less interpretable than traditional semantic maps, they seem to encode more information and capture accurate details.
% We observe that the learned latent layouts improves the performance compared to predicting semantic segmentations for our task.

% The authors of SPADE~\cite{park2019SPADE} originally utilized semantic segmentations as the conditional input to their generator.
% Semantic labeling provides a dense spatial map aligns with the target image, which is suitable for computing per-pixel de-modulation parameters.
% However, an interesting question is whether we can learn a dense latent spatial representation that better suits the image synthesis task at hand.
% To do that, we remove the cross-entropy loss supervision to give the layout predictor, $\glayout$, the freedom to diverge from predicting traditional semantic segmentations to learning other latent representations that better optimize the few-shot novel view synthesis objective.
% The layout predictor is only supervised by the training objective of Eqn.~\ref{eqn:objective}.

%--------------------------------
\begin{figure*}[t!]
    \centering
    \includegraphics[width=0.88\linewidth]{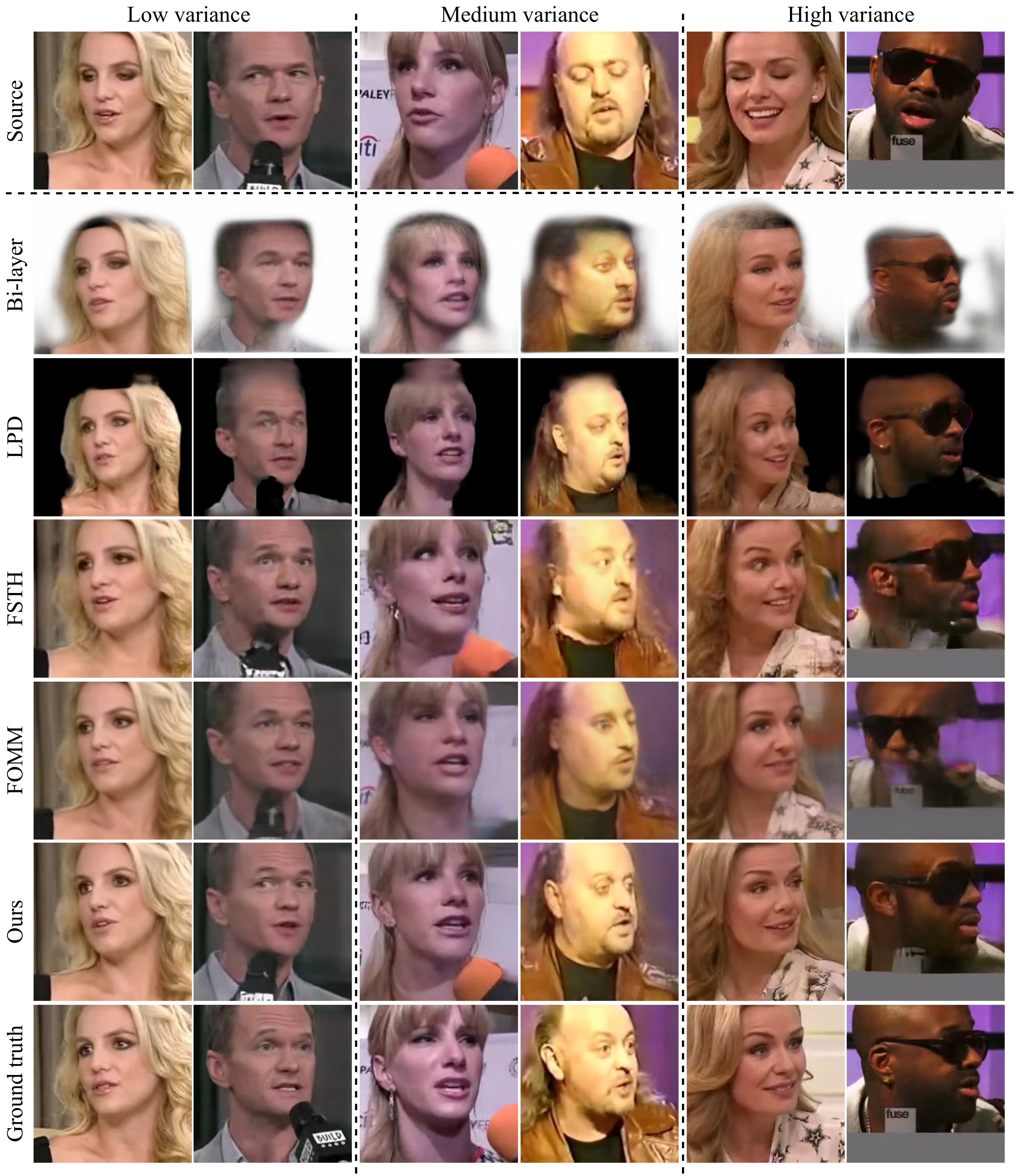}
    \caption{Qualitative comparison in the single-shot setting. We show three sets of examples representing low, medium and high variance between the source and target poses. Our method is more robust to pose variations than the baselines.}
    \vspace{-0.2cm}
    \label{fig:qual_cmp}
\end{figure*}

%--------------------------------

% ------------------------------------------------------------------------------------
\subsection{Subject fine-tuning}
\label{sec:finetuning}
Training our full pipeline learns a powerful subject-agnostic model that produces high quality and identity-preserving synthesis.
Optionally, we can learn a personalized head avatar to further refine the results for a given subject.
To do this, we follow~\cite{zakharov2019few,burkov2020neural,Zakharov20fast} and fine-tune the subject-agnostic model (also called \emph{meta-learned} model) using the few-shot inputs of the source identity.
Specifically, we compute the layout and style embeddings $\{\zlayout, \zstyle\}$ and fine-tune the weights of the layout and image generators $\{\glayout, \gstyle\}$, as well as the discriminator, $D$, by reconstructing the set of few-shot inputs, and optimizing the same training objective of Eqn.~\ref{eqn:objective}.
We observe that subject fine-tuning restores high-frequency components and improves background reconstruction when compared to the meta-learned outputs.
%
% --------------------------------------
% We follow~\cite{zakharov2019few,burkov2020neural,Zakharov20fast} by introducing a subject-specific fine-tuning phase that aims at improving the network initialization for an unseen target subject at test time.
% %
% Given the few-shot inputs of a target subject, we compute the layout and style embeddings $\zlayout, \zstyle$ only once, and we fine-tune the weights of the layout and image generators $\glayout, \gstyle$, as well as the discriminator, $D$, by reconstructing the set of few-shot inputs. The fine-tuning objective is the same as Eqn~\ref{eqn:objective} of the training objective, but we only fine-tune the $\glayout, \gstyle$, and $D$ sub-networks while the encoders $E^l, E^s$ are kept fixed.
% %
% Subject fine-tuning restores high-frequency details, improves identity preservation, and restores background details when compared to the meta-learned outputs.
% %
% The fine-tuning phase is carried out only once per target subject and runs for a few steps.
% This typically takes an order of seconds to a minute depending on the number of input few-shots.

\section{Experimental evaluation}
\label{sec:experiments}

% ----------------------------------------
% \input{figures_tex/qual_cmp}
\begin{table}[t!]
  \caption{Quantitative comparison in the single-shot setting.}
  \label{table:quantitative}
  \centering
   \renewcommand{\arraystretch}{1.2}
   \renewcommand{\tabcolsep}{3pt} % {1.05mm} %{1.5mm} % (1.5mm = 4.252 pts)
  \resizebox{1.0\linewidth}{!}{
%   \begin{footnotesize}
  \footnotesize
      \begin{tabular}{@{}lcccccc@{}}
    %   \begin{tabular}{@{}lcccccc@{}}
        \toprule
         Method & PSNR$\uparrow$ & SSIM$\uparrow$ & LPIPS$\downarrow$ & \idsim $\uparrow$ & \nmke $\downarrow$ & FID$\downarrow$  \\ 
         \midrule % \cline{2-13}%\cmidrule(3){2-13}
         X2Face~\cite{wiles2018x2face} & 15.50 & 0.466 & 0.346 & 0.691 & 0.333 & 98.58 \\
         Bi-layer~\cite{Zakharov20fast} & -- & -- & -- & 0.721 & 0.236 & 130.58 \\
         FSTH~\cite{zakharov2019few} & 16.92 & 0.597 & 0.263 & 0.836 & \underline{0.049} & 53.07 \\
         LPD~\cite{burkov2020neural} & -- & -- & -- & 0.837 & 0.070 & \underline{48.48} \\
         FOMM~\cite{Siarohin_2019_NeurIPS} & \textbf{18.20} & \textbf{0.635} & \underline{0.236} & \underline{0.869} & 0.061 & 56.10 \\
        %  Ours - meta & 17.27 & 0.598 & 0.241 & 0.869 & ?? & 48.11 \\
         Ours & \underline{17.37} &	\underline{0.605} &	\textbf{0.232} & \textbf{0.886} & \textbf{0.041} & \textbf{45.69} \\
        \bottomrule
      \end{tabular}
%   \end{footnotesize}
  }
% \vspace{-0.17in}
\end{table}

% % ------------------------------------------------------------------------------

% ----------------------------------------
\paragraph{Implementation details.} Please, refer to the supp.~material for networks architecture, hyper-parameters and training details.
Our code will be publicly released.
%% -----------------------------------------------------------------------------
\paragraph{Dataset.}
We perform our evaluation using the VoxCeleb~\cite{Chung18b} dataset, which is a large-scale in-the-wild video dataset.
The train set contains over a million clips from 145,569 videos of 5,994 different identities.
The test set contains new identities that are not part of the training.
We use the test subset released by Zakharov \etal~\cite{zakharov2019few}, which contains a total of 1,600 frames from videos of 50 subjects.
For self-reenactment scenarios, the input few-shots and the driving sequence do not overlap.
We obtain the facial landmarks for sampled frames using an off-the-shelf facial landmarks detector~\cite{bulat2017far}.

%% -----------------------------------------------------------------------------
\paragraph{Baselines.}
% We compare our model against two groups of baselines.
% The first group includes FOMM~\cite{Siarohin_2019_NeurIPS} and Bi-layer~\cite{Zakharov20fast} which operate only in a single-shot setting.
% The second group accepts few-shot inputs and includes X2Face~\cite{wiles2018x2face}, FSTH~\cite{zakharov2019few}, and Latent Pose Descriptors (LPD)~\cite{burkov2020neural}.
%
% We compare our model to single-shot baselines (FOMM~\cite{Siarohin_2019_NeurIPS} and Bi-layer~\cite{Zakharov20fast}), and few-shot baselines (X2Face~\cite{wiles2018x2face}, FSTH~\cite{zakharov2019few}, and Latent Pose Descriptors (LPD)~\cite{burkov2020neural}).
%
% We compare our model to both single-shot and few-shot state-of-the-art. 
% Single-shot baselines include FOMM~\cite{Siarohin_2019_NeurIPS} and Bi-layer~\cite{Zakharov20fast}, and few-shot baselines include X2Face~\cite{wiles2018x2face}, FSTH~\cite{zakharov2019few}, and Latent Pose Descriptors (LPD)~\cite{burkov2020neural}.
% %
% We first compare to all baselines in a single-shot setting, and further compare our model against few-shot baselines in a multi-shot setting.
% %
We compare our method to the following baselines: 
X2Face~\cite{wiles2018x2face}, FSTH~\cite{zakharov2019few}, FOMM~\cite{Siarohin_2019_NeurIPS}, Latent Pose Descriptor (LPD)~\cite{burkov2020neural}, and Bi-layer~\cite{Zakharov20fast}.
We use the released pre-trained models provided by the authors for all baselines, except for FSTH~\cite{zakharov2019few} where we use the authors' provided outputs, as their code and models were not released.
Since some baselines only accept single-shot inputs (\eg FOMM and Bi-layer), we divide our evaluation into a single-shot setting, where we compare to all the baselines, and a multi-shot setting, where we only compare against the few-shot baselines.
Since the LPD~\cite{burkov2020neural} and Bi-layer~\cite{Zakharov20fast} baselines do not predict the background and re-crop the input/output frames, we subtract the background and compare with their corresponding cropped ground truths for quantitative analysis.
We also exclude those two baselines from frame reconstruction evaluation since their output does not align with the rest of the methods.

%% -----------------------------------------------------------------------------
\paragraph{Metrics.}
We evaluate all models along five axes.
\begin{itemize}[leftmargin=*]
    \item Reconstruction fidelity using the peak signal-to-noise ratio (\emph{PSNR}) and structural similarity (\emph{SSIM})~\cite{wang2004image} metrics.
    \item Perceptual similarity between the output and the ground truth  using the \emph{AlexNet}-based \emph{LPIPS} metric~\cite{zhang2018unreasonable}.
    \item Identity preservation (\emph{\idsim}) using the cosine similarity between face embeddings from a face recognition network~\cite{parkhi2015deep}.
    \item Normalized Mean Keypoint Error (\emph{\nmke}), which measures the pose error between the synthesized and ground truth images as computed in~\cite{burkov2020neural,Zakharov20fast}.
    \item Perceptual quality of the output using the Frechet-Inception Distance (\emph{FID}) metric~\cite{heusel2017gans}.
\end{itemize}
%% -----------------------------------------------------------------------------

\subsection{Single-shot comparative evaluation}
\label{sec:exp_single_shot}
Table~\ref{table:quantitative} shows a quantitative comparison with the baselines in the single-shot setting.
Our method outperforms all baselines in perceptual reconstruction (LPIPS), identity preservation (\idsim), pose matching (\nmke) and visual quality (FID).
However, FOMM scores better in the standard reconstruction metrics (PSNR and SSIM).
We argue this is intrinsic to their method due to its warping-based nature, which accurately captures the background and other static regions, and thus gives low reconstruction error even in the presence of clear artifacts.
Furthermore, while FOMM cannot utilize more input frames to its advantage, our method's performance improves with multi-shot inputs to significantly surpass FOMM in all metrics (see supp.~material for the quantitative numbers).

Figure~\ref{fig:qual_cmp} shows qualitative results from three groups representing low, medium and high variance between the input and target poses.
We observe that all methods perform well when the target pose is similar to that of the input shot.
LPD produces sharp results within the low-medium pose variation, but shows blurry artifacts within the face and eyes in the case of high pose variance.
FSTH shows a clear identity gap.
FOMM accurately matches the background and shows highly realistic results when the pose variance is low, but shows a clear identity gap and visible artifacts when the target pose is far from the source image.
Our method is more robust against pose variation, yielding realistic results while preserving the source identity.
%% -----------------------------------------------------------------------------
% ----------------------------------------
\begin{figure}[t!]
    \centering
    \includegraphics[width=0.92\linewidth]{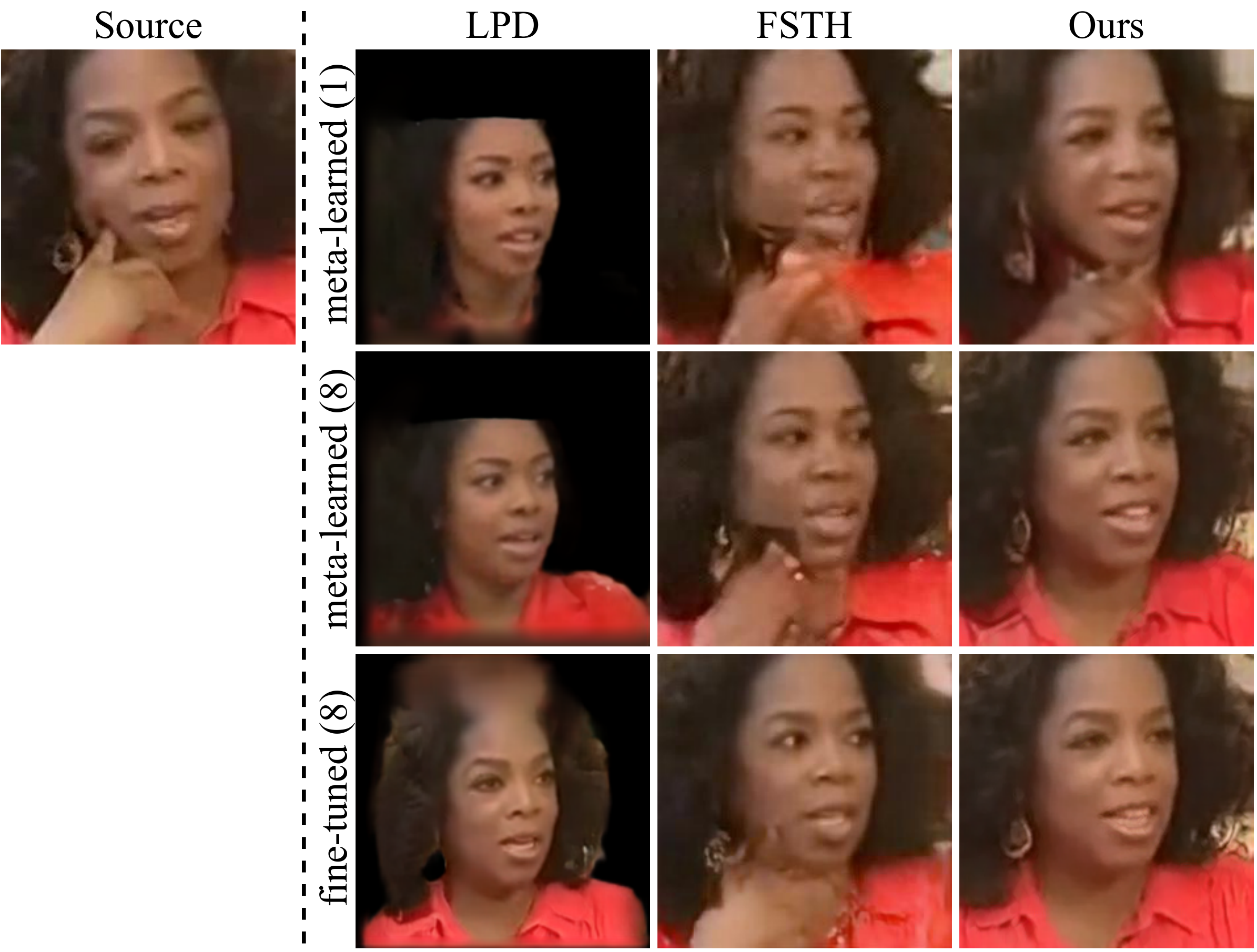}
    \caption{A qualitative comparison showing the effect of increasing the K-shot inputs and applying subject fine-tuning.}
    \vspace{-0.2cm}
    \label{fig:var_k_and_ft}
\end{figure}

% ----------------------------------------
\begin{figure*}[ht!]
    \centering
    \includegraphics[width=0.73\linewidth]{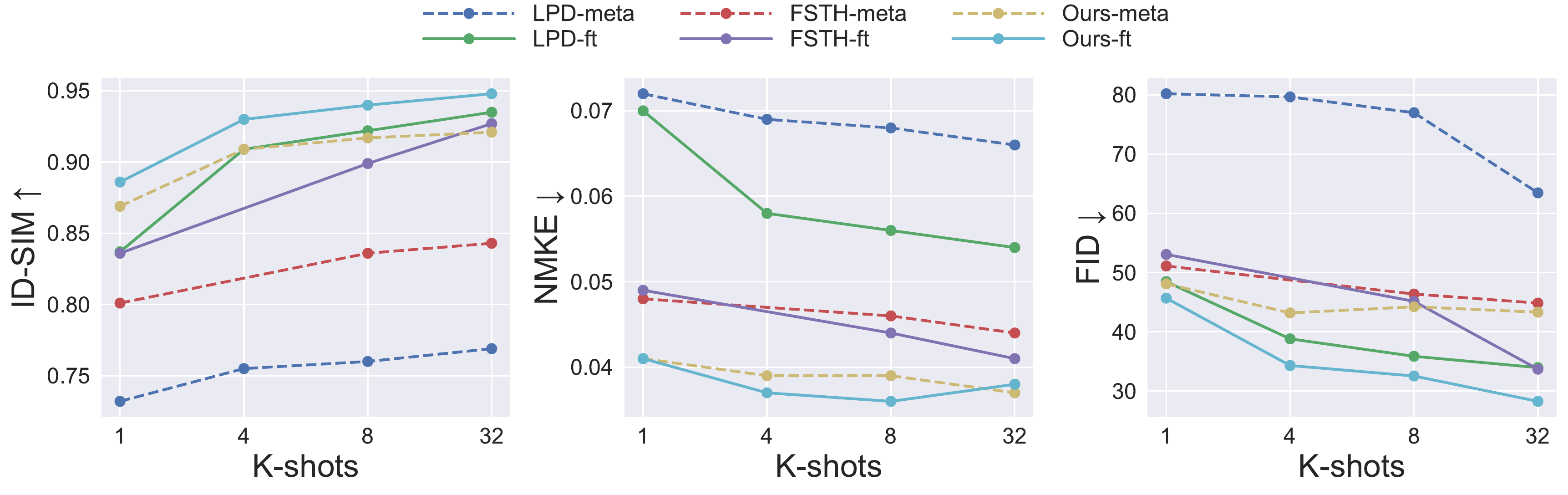}
    \caption{Quantitative comparison with the few-shot baselines, showing the effect of both increasing the K-shot inputs and subject-specific fine-tuning. Dotted and solid lines represent the meta-learned and fine-tuned models respectively.}
    % \vspace{-0.1cm}
    \label{fig:var_k_and_ft_graphs}
\end{figure*}

% ----------------------------------------
\begin{figure*}[ht]
    \vspace{-0.3cm}
    \centering
    \includegraphics[width=0.95\linewidth]{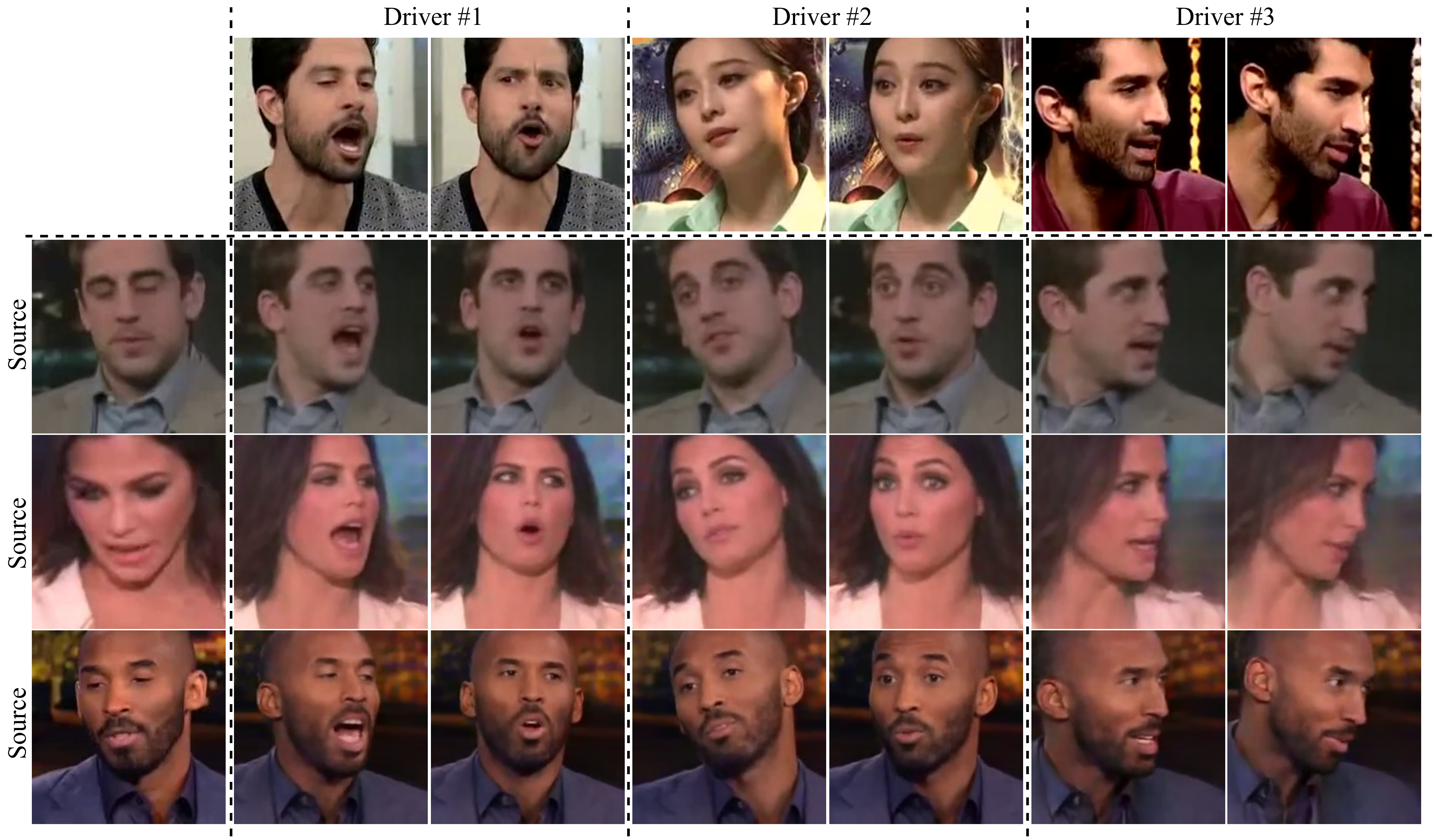}
    \caption{Cross-subject reenactment with different driving identities. Results are shown for our \emph{meta-learned} model without any fine-tuning, and using 32-shot inputs.}
    \vspace{-0.4cm}
    \label{fig:reenactment}
\end{figure*}

% ----------------------------------------

% ----------------------------------------
\begin{figure*}[ht]
    \centering
    \includegraphics[width=0.75\linewidth]{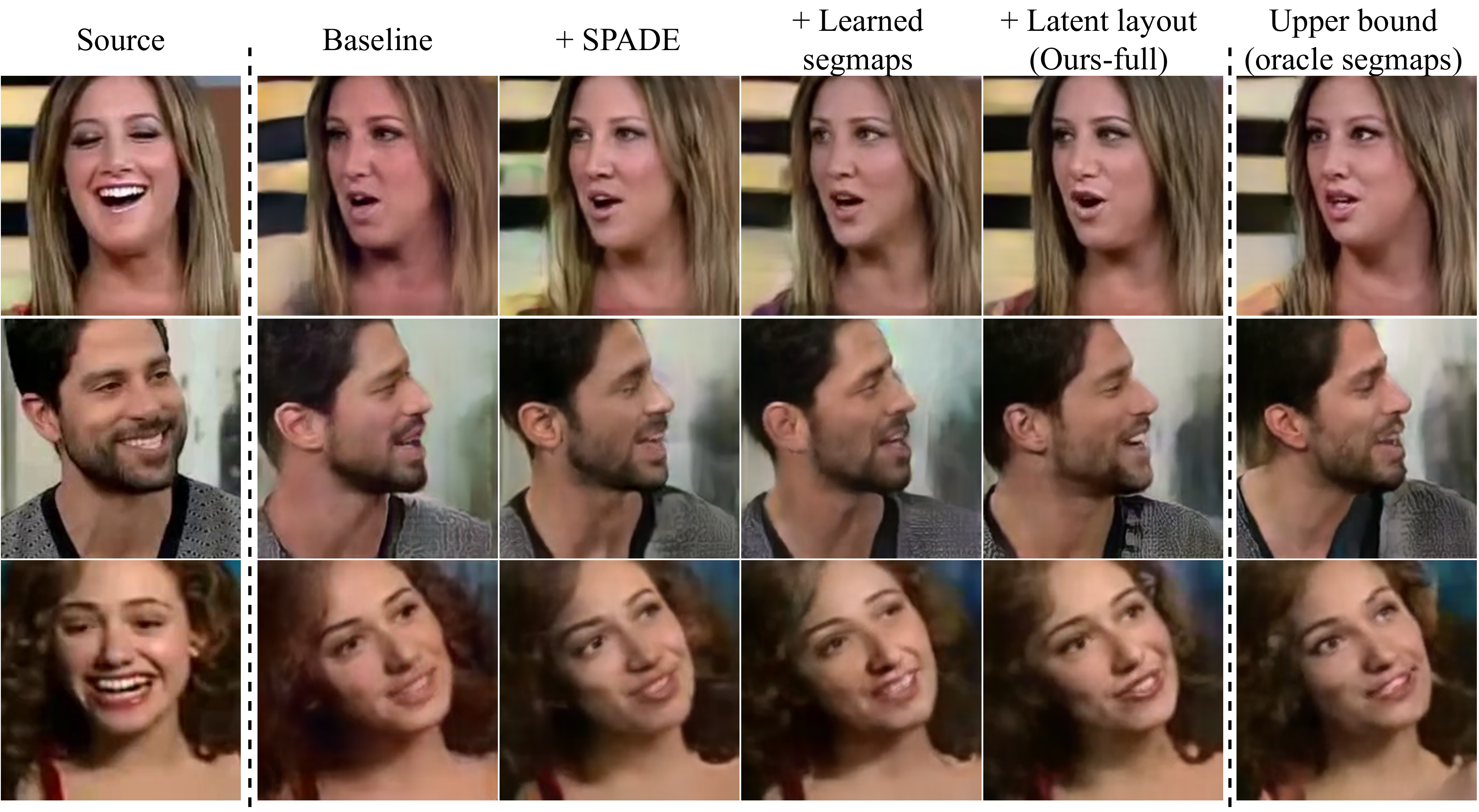}
    \caption{Examples from the ablation study. Results shown are for the meta-learned models with a single-shot input (source).}
    \vspace{-0.4cm}
    \label{fig:ablation}
\end{figure*}

% ----------------------------------------

\subsection{Multi-shot comparative evaluation}
\label{sec:exp_multi_shot}
% Here we compare our performance in the multi-shot setting against FSTH~\cite{zakharov2019few} and LPD~\cite{burkov2020neural}. 
% We evaluate our method and the baselines using K-shot inputs for K $\in {1, 4, 8, 32}$.
% We report results in two settings, namely meta-learned and fine-tuned.
% In the meta-learned setup, we feed-forward the inputs through the network to obtain the output.
% On the other hand, the fine-tuned setting fine-tunes the model using the few-shot inputs to learn a personalized model.
% -------------------------------------------------------------
% Here we evaluate our method in the multi-shot setting.
Here, we focus on the effect of increasing the number of $K$-shot inputs, and the effect of subject-specific fine-tuning using the $K$-shot inputs.
% We plot the results in Figure~\ref{fig:var_k_and_ft_graphs}.
Figure~\ref{fig:var_k_and_ft_graphs} plots the \idsim, {\nmke} and FID performance metrics as we increase the number of $K$-shots.
%  for identity preservation (\idsim), pose error (\nmke) and visual quality (FID).
%
% the more data -- the longer the finetuning -- the better the results -- but the closer we are to subject-specific models.
%
We observe that the pose reconstruction performance (\nmke) is mainly dictated by the approach itself, rather than the number of $K$-shots or whether the models are fine-tuned or not.
For example, the \emph{meta-learning} performance of FSTH with $K=1$ is better than the \emph{fine-tuned} LPD model with $K=32$.
Similarly, the \emph{single-shot meta-learning} performance of our method is better than the \emph{fine-tuned} baselines at $K=32$.

For the \idsim \ and FID metrics, the \emph{meta-learning} performance of our model is not only superior to that of the baselines, but it is also on-par with the \emph{fine-tuned} baselines for $K \le 8$. However, as $K$ is increased to 32, the \emph{fine-tuned} baselines eventually outperform our \emph{meta-learned} model.
Another very important advantage to our approach is that it achieves better performance with significantly less data.
For example, fine-tuning our model with just $K=4$ outperforms the fine-tuned baselines at $K=32$.
Since fine-tuning on more data requires more training iterations and thus more time, our method spends much less time fine-tuning on fewer data samples, and yet achieves similar or better results.
We observe similar behavior with other metrics (PSNR, SSIM and LPIPS).
Please, refer to the supp.~material for full results.

Figure~\ref{fig:var_k_and_ft} visualizes the effect of both increasing $K$ and subject fine-tuning.
Our method preserves the source identity without any fine-tuning, even with a single-shot input.
% Our performance gradually improves as we increase $K$ and after fine-tuning.
On the other hand, the baselines only restore the source identity after the subject-specific fine-tuning.
Our method also shows the most improvement, in terms of realism and better identity match, when increasing the number of $K$-shot inputs.
For example, our method successfully filters out the subject hand occluding the face in the single-shot input.
% Increasing the number of K-shots helps cancel out transient parts (\eg the subject hand occluding the face) and results in higher realism, with our method showing the most improvement.

% Furthermore, increasing the K-shot inputs from 1 to 8 improves the realism and identity preservation of our method.
% highlights the qualitative difference with the baselines  between our method and the baselines as we increase $K$ and fine-tune.
%% -----------------------------------------------------------------------------

\subsection{Cross-subject reenactment}
\label{sec:reenactment}
Cross-subject reenactment poses a challenge, especially for landmark-driven approaches.
The shape difference between facial landmarks of the source and driver identities could lead to a noticeable identity gap in the reenacted results.
The intermediate spatial representation learned by our method helps reduce this problem and leads to good identity preservation of the source subject regardless of the driver identity.
Figure~\ref{fig:reenactment} shows sample reenactment results using different driver identities.
% The driving sequences cover views of both the left and right sides of face and span challenging facial expressions.
To demonstrate the effectiveness of our disentangled representation, we avoid any subject fine-tuning and show the results of our meta-learned model with 32-shot inputs.
The source identity is well-preserved among challenging facial expressions and different views covering both the left and right sides of the face.
%% -----------------------------------------------------------------------------

% ----------------------------------------
% \input{figures_tex/ablation}
% PSNR & LPIPs metrics only
\begin{table}[ht!]
% \vspace{-0.1in}
  \caption{Ablation study of our approach. +SPADE replaces the UNet generator with SPADE. +Learned seg.~maps conditions the generator on learned segmentations. +Latent layout learns a latent spatial representation. The upper bound gets to cheat and uses the ground truth segmentations.}
  \label{table:ablation}
%   \vspace{-0.25cm}
  \centering
  \footnotesize
  \renewcommand{\arraystretch}{1.2}
  \renewcommand{\tabcolsep}{3pt}
  \resizebox{1.0\linewidth}{!}{
  \begin{footnotesize}
      \begin{tabular}{@{}lcccccc@{}}
        \toprule
         Method & PSNR$\uparrow$ & SSIM$\uparrow$ & LPIPS$\downarrow$ & \idsim $\uparrow$ & \nmke $\downarrow$ & FID$\downarrow$  \\ 
         \midrule % \cline{2-13}%\cmidrule(3){2-13}
         Baseline & \underline{17.00} & 0.574 & 0.274 & \underline{0.837} & 0.044 & 67.19 \\
         + SPADE & 16.94 & 0.575 & 0.268 & 0.834 & 0.043 & \underline{56.00} \\
         + learned seg.~maps & 16.94 & \underline{0.578} & \underline{0.265} & 0.828 & \textbf{0.042} & 62.78 \\
         + latent layout (\textbf{ours}) & \textbf{17.22} & \textbf{0.592} & \textbf{0.247} & \textbf{0.860} & \underline{0.042} & \textbf{54.40} \\
         \midrule
         Upper bound & 18.21 & 0.629 & 0.219 & 0.867 & 0.039 & 48.06 \\
        \bottomrule
      \end{tabular}
  \end{footnotesize}
  }
\vspace{-0.2cm}
\end{table}

% % ------------------------------------------------------------------------------

% ----------------------------------------

\subsection{Ablation study}
\label{sec:ablation}
We evaluate the contribution of different components of our proposed approach.
All ablation experiments are trained with the same hyper-parameters and for the same number of epochs, and are evaluated in the \emph{single-shot} setting with \emph{no fine-tuning}.
We report the results in Table~\ref{table:ablation}.
The baseline model has the same setup as FSTH~\cite{zakharov2019few}, where a UNet generator with AdaIN layers~\cite{huang2017arbitrary} translates the input landmarks into the target image.
Next, we replace the UNet architecture with a SPADE generator~\cite{park2019SPADE} conditioned on the facial landmarks (+SPADE).
This improved the FID, but other metrics remained around the same.
We hypothesize this is due to using sparse landmarks as the spatial input, while SPADE needs dense spatial inputs to generate the per-pixel denormalization parameters.
To validate our hypothesis, we conducted an experiment as an \emph{upper bound}, where we get to cheat and segment the ground truth target image using an off-the-shelf face segmentation network~\cite{CelebAMask-HQ} (\ie oracle), and we use these oracle segmentations as the spatial input to SPADE.
Even though the oracle segmentations are noisy (\eg Figure~\ref{fig:layout}), this still resulted in a significant boost in all metrics, proving that the SPADE generator could benefit from dense spatial inputs.
Therefore, we trained a layout prediction network to predict a plausible semantic segmentation for the target pose (+Learned seg.~maps).
This surprisingly produced mixed results and even caused a drop in the {\idsim} and FID scores.
We posit this is because the noisy oracle segmentations do not provide a consistent supervisory signal, which causes the learned segmentations to miss important shape cues (\eg the correct face shape), as well as overfit common errors in the oracle segmentations as the training progresses.
Finally, removing the supervision on the predicted layouts and learning a latent spatial representations (+Latent layouts) resulted in a reasonable performance improvement over all metrics.
We also show a qualitative comparison for the ablation study in Figure~\ref{fig:ablation}.
We observe that the qualitative results of the upper bound experiment (using the oracle/ground-truth segmentation) exhibits artifacts caused by errors in the oracle segmentation.
% Our method (+Latent layout) shows a more realistic output with no clear artifacts, despite . 
The results of our method with the learned latent layouts looks qualitatively better, with no clear artifacts, despite having worse quantitative metrics than the upper bound experiment.
%
% We observe that the qualitative results of the upper bound experiment (using the oracle/ground-truth segmentation) is not as good compared to our full method (+Latent layout). 
% Even though using the oracle/ground-truth segmentations gives an upper bound on the quantitative metrics, the qualitative results still shows clear artifacts due to the noise in the oracle segmentations. 
% The results of our method with the learned latent layouts looks qualitative better than the upper bound experiment, despite what the quantitative metrics show.
%% -----------------------------------------------------------------------------

% \subsection{Pose variation}
% \label{sec:pos_var}

%% -----------------------------------------------------------------------------

\section{Conclusion}
\label{sec:conclusion}
% \paragraph{Conclusion.}
We proposed a novel approach for talking-head synthesis.
Our model learns a novel latent spatial representation that proves effective for our task.
We improve the performance of both subject-agnostic models, as well as subject-finetuned models while requiring significantly less data samples.
% Our subject-agnostic model achie
%
The learned latent spatial representation helps provide robustness against a wide range of poses and expressions, and results in better identity preservation, especially for the cross-subject reenactment scenarios.

% \bigskip
\medskip
% \smallskip
{\footnotesize \noindent\textbf{Acknowledgements.}
% {\noindent\textbf{Acknowledgements.}
% We would like to thank the members of the Perception and Intelligence (PI) Lab for their helpful feedback.
% This project was partially funded by  DARPA SemaFor (HR001119S0085), DARPA MediFor (FA87501620191) and DARPA SAIL-ON (W911NF2020009) programs.}
This project was partially funded by the DARPA SemaFor (HR001119S0085) and DARPA SAIL-ON (W911NF2020009) programs.}

% % \tableofcontents
% % \pagebreak
% \newpage
% \appendix
% \input{appendix.tex}

%%%%%%%%% References

{\small
\bibliographystyle{ieee_fullname}
\bibliography{references}
}

% % \pagebreak
\newpage
\appendix
\section{Appendix}
\label{sec:appendix}
\subsection{Implementation details}
\label{sec:implementation}
\paragraph{Dataset pre-processing.}
The released VoxCeleb2 dataset~\cite{Chung18b} contains pre-processed videos to have a center crop around the face.
We uniformly sample 10 frames from each video and obtain the facial landmarks using an off-the-shelf facial landmarks detector~\cite{bulat2017far}.
Once the landmarks are obtained we use the same procedure as~\cite{zakharov2019few} to connect the facial landmarks to obtain contours for different face parts (\eg eyes, nose, lips \dots etc.~).
We observe that the facial landmarks extraction fails for a small fraction of videos, which we opted to ignore.
We also segment each frame using the face parsing tool provided by~\cite{CelebAMask-HQ} to obtain the oracle segmentation maps for pre-training the layout prediction network.
The face parsing network performs poorly on VoxCeleb2 frames due to the domain gap, in terms of image resolution and the distribution of head poses, between the datasets used to train the face parsing network~\cite{CelebAMask-HQ}, and the cropped VoxCeleb2 videos.
We observe that the face segmentation network~\cite{CelebAMask-HQ} better captures different details at different resolutions.
For example the segmentation result at the original VoxCeleb2 resolution better captures larger regions like the hair, neck and clothes.
On the other hand, upsampling the frame to the resolution used for training the segmentation network~\cite{CelebAMask-HQ} gives better segmentation results for the finer and smaller regions like the nose, eyes, mouth, and ears.
So, to improve the oracle segmentations, we segment each frame twice at 256x256 and 512x512 resolutions and merge the coarse and fine semantic classes from both results.
% We use a heuristic based approach to merge these two maps finally outputting a single map.
%
\paragraph{Encoder networks.} We use a resnet encoder for both the layout and style encoders $\{ \elayout, \estyle \}$.
The encoder architecture has 5 downsampling blocks, followed by a fully connected layer that generates a 512-dimensional latent code.
The architecture for the residual blocks is borrowed from~\cite{brock2018large}, with replacing \emph{average-pooling} with \emph{blur-pooling}.
We use 32 feature maps at the first encoder layer and double this number after each downsampling block with a maximum of 512 feature maps.
We follow~\cite{zakharov2019few} and concatenate the facial landmarks to the few-shot RGB images before feeding them to the encoder.
\paragraph{Layout generator.} We use a traditional UNet architecture~\cite{ronneberger2015u} with residual blocks.
The residual blocks are borrowed from~\cite{brock2018large} with replacing \emph{BatchNorm} with \emph{Instance Norm} and applying adaptive instance normalization (\emph{AdaIN})~\cite{huang2017arbitrary}.
The smallest and largest number of feature maps are 32 and 512 respectively, and we use \emph{blur-pool} and \emph{bilinear} upsampling in the downsampling and upsampling blocks respectively.
\paragraph{Image generator.} We use a SPADE generator architecture~\cite{park2019SPADE} with replacing \emph{BatchNorm} with \emph{Instance Norm}. We also use 32 feature maps at the last generator layer and 64 feature maps in each SPADE block, compared to 64 and 128 feature maps respectively in the original architecture~\cite{park2019SPADE}.
The input to each SPADE block is the concatenation of the predicted layout map and the facial landmarks.
\paragraph{Discriminator.} We borrow the architecture of the discriminator network
from~\cite{karras2020analyzing}, with reducing the smallest number of feature maps from 64 to 32. 
We also use a non-saturating logistic loss with gradient
penalty~\cite{mescheder2018training}.
% <<<<<<<<<<<<<<<<<<<<<<<<<<<<<<<<<<<<<
\begin{figure}[t!]
    \centering
    \includegraphics[width=0.9\linewidth]{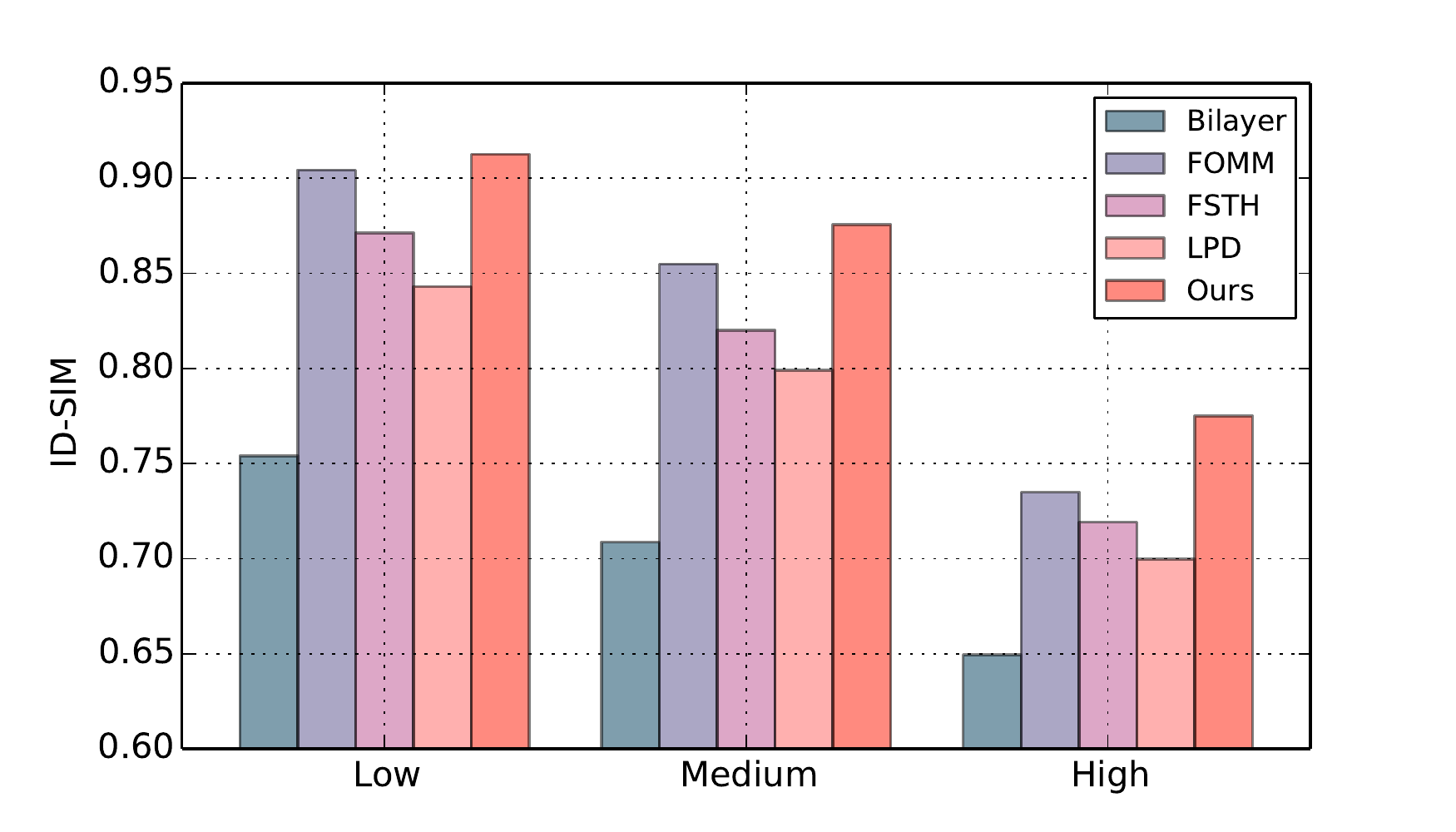}
    \caption{Identity similarity metric (ID-SIM) for the single-shot setting across three test subsets representing low, medium and high variance between the source and target poses.
    The performance gap widens in our favor as the pose variance increases.}
    % Our method is more robust to pose variations than the baselines.}
    % \vspace{-0.2cm}
    \label{fig:pose_var}
\end{figure}

% >>>>>>>>>>>>>>>>>>>>>>>>>>>>>>>>>>>>>
%
\paragraph{Training.} We follow~\cite{karras2017progressive} and use equalized learning rate in all of our networks.
We pre-train the layout prediction network for 2 epochs, followed by training the full pipeline for 8 epochs. 
Our best model was left to train for an extra 5 epochs, which mainly improves the FID score, while slightly improving the other quantitative metrics as well.
We use an \emph{Adam} optimizer~\cite{kingma2014adam} with $\beta_1=0, \beta_2=0.999$, and a learning rate of $0.001$ for all networks.
We linearly decay the learning rate by a factor of $100$ during the last epoch.
% TODO: loss weights
% TODO: subject fine-tuning details.
%
For more implementation and training details, we will release the code and training scripts at \url{http://www.cs.umd.edu/~mmeshry/projects/lsr/}.
% --------------------------------------------------------------

\subsection{Robustness to pose variation}
\label{sec:pose_var}
% - Bar plot + more single-shot comparison.\\
% - LPD does well but it requires subject fine-tuning.\\
Here we investigate the robustness of different methods against the pose variation between the source and the target images.
First, we cluster the test set into low, medium and high pose variance based on the mean normalized keypoint difference (\nmke) between the source and target ground truth images.
Then we compute the identity similarity metric (\idsim) per each cluster for the single-shot setting and report the results in figure~\ref{fig:pose_var}.
The performance gap between our method and the baselines widens as the pose variance increases, indicating that our method has better robustness against pose variation.
Note that we report the results only for the single-shot setting, where the performance gap with the FOMM baseline~\cite{Siarohin_2019_NeurIPS} is close.
However, our method significantly outperforms FOMM in the multi-shot setting, as we show in Section~\ref{sec:fomm_cmp}.
% while our method generally shows a significantly better performance gap in the multi-shot setting (\eg $K=4$), as .
% --------------------------------------------------------------

% <<<<<<<<<<<<<<<<<<<<<<<<<<<<
\begin{figure}[t!]
    \centering
    \includegraphics[width=0.88\linewidth]{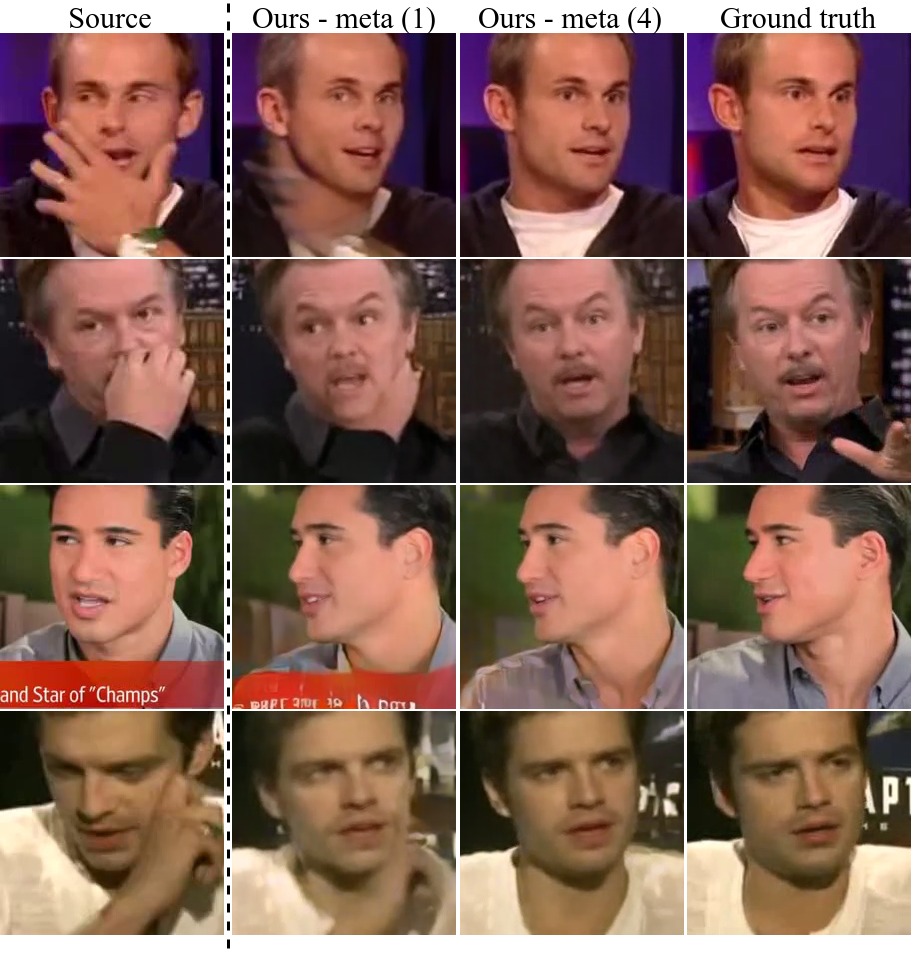}
    \caption{Averaging latent codes from multi-shot inputs successfully filters out transient occluders and maintains only the desired information for novel view head synthesis.}
    % \vspace{-0.2cm}
    \label{fig:latent_avg}
\end{figure}

% >>>>>>>>>>>>>>>>>>>>>>>>>>>>
\subsection{Effect of latent averaging}
\label{sec:latent_avg}
Given K-shot inputs, we follow~\cite{zakharov2019few} and obtain a single layout and style latents \{$\zlayout, \zstyle$\} by averaging the K layout and style latents computed from the inputs respectively.
We observe that averaging the K latents cancels out view-specific information and transient occluders, and successfully maintains
the implicit 3D information needed for novel view head synthesis.
Figure~\ref{fig:latent_avg} shows some examples that highlight this effect.
The single-shot source images show some transient occluders like the subjects' hand or news bar, which in turn corrupts our single-shot output.
However, increasing the inputs to four frames successfully filters out the transient occluders and results in clean outputs.
% --------------------------------------------------------------

\subsection{More comparative results}
\label{sec:more_cmp}
% <<<<<<<<<<<<<<<<<<<<<<<<<
% PSNR & LPIPs metrics only
\begin{table}[t!]
% \vspace{-0.1in}
  \caption{Comparison with the FOMM baseline~\cite{Siarohin_2019_NeurIPS}. While FOMM cannot benefit from multiple input frames, our method shows a significant improvement over FOMM with as few as 4-shot inputs.}
  \label{table:fomm_cmp}
%   \vspace{-0.25cm}
  \centering
  \footnotesize
  \renewcommand{\arraystretch}{1.2}
  \renewcommand{\tabcolsep}{3pt}
  \resizebox{1.0\linewidth}{!}{
  \begin{footnotesize}
      \begin{tabular}{@{}lcccccc@{}}
        \toprule
         Method & PSNR$\uparrow$ & SSIM$\uparrow$ & LPIPS$\downarrow$ & \idsim $\uparrow$ & \nmke $\downarrow$ & FID$\downarrow$  \\ 
         \midrule % \cline{2-13}%\cmidrule(3){2-13}
         FOMM~\cite{Siarohin_2019_NeurIPS} &  \textbf{18.20} & \textbf{0.635} & 0.236 & 0.869 & 0.061 & 56.10 \\
         Ours-meta (K=1) & 17.27 & 0.598 & 0.241 & 0.869 & 0.041 & 48.11 \\
         Ours-ft (K=1) & 17.37 & 0.605 & \textbf{0.232} & \textbf{0.886} & \textbf{0.041} & \textbf{45.69} \\
         \midrule
         Ours-meta (K=4) & 18.90 & 0.638 & 0.192 & 0.909 & 0.039 & 43.19 \\
         Ours-ft (K=4) & \textbf{19.33} & \textbf{0.661} & \textbf{0.171} & \textbf{0.930} & \textbf{0.037} & \textbf{34.31} \\
        \bottomrule
      \end{tabular}
  \end{footnotesize}
  }
\end{table}

% % ------------------------------------------------------------------------------

% >>>>>>>>>>>>>>>>>>>>>>>>>
\subsubsection{Comparison with FOMM}
\label{sec:fomm_cmp}
The FOMM baseline~\cite{Siarohin_2019_NeurIPS} accurately reconstructs the background and other static regions due to its warping-based nature.
Therefore, it achieves lower reconstruction error (PSNR and SSIM) than our approach in the single-shot setting, even if their output contains clear artifacts in the face area.
However, one limitation to FOMM is that it cannot utilize more input frames to its advantage.
On the other hand, Table~\ref{table:fomm_cmp} shows that our approach benefits from as few as four input frames to outperform FOMM, even in the meta-learned mode.
Subject fine-tuning further improves our performance to outperform FOMM by a wide margin in all metrics.

% --------------------------------------------------------------
% <<<<<<<<<<<<<<<<<<<<<<<<<<<<<
\begin{table*}
    \centering
    % \small
    \footnotesize
    \caption{Detailed quantitative comparison with the few-shot baselines, showing the effect of both increasing the K-shot inputs and subject-specific fine-tuning.}
    \label{table:var_k_and_ft}
    \begin{tabular}{@{}clcccccccccccc@{}}
    \toprule
    \multirow{2}{*}{K} & \multirow{2}{*}{Method} & \multicolumn{6}{c}{No Subject Fine-tuning} & \multicolumn{6}{c}{Subject Fine-tuned}\\
    \cmidrule[\cmidrulewidth](l){3-8}
    \cmidrule[\cmidrulewidth](l){9-14}
    & & PSNR$\uparrow$ & SSIM$\uparrow$ & LPIPS$\downarrow$ & \idsim $\uparrow$ & \nmke $\downarrow$ & FID$\downarrow$ & PSNR$\uparrow$ & SSIM$\uparrow$ & LPIPS$\downarrow$ & \idsim $\uparrow$ & \nmke $\downarrow$ & FID$\downarrow$ \\
    \midrule
    \multirow{3}{*}{1} & FSTH & 16.80 & 0.570 & 0.259 & 0.801 & 0.048 & 51.12 & 16.92 & 0.597 & 0.263 & 0.836 & 0.049 & 53.07 \\
    & LPD & -- & -- & -- & 0.732 & 0.072 & 80.20 & -- & -- & -- & 0.837 & 0.070 & 48.48 \\
    & Ours & \textbf{17.27} & \textbf{0.598} & \textbf{0.241} & \textbf{0.869} & \textbf{0.041} & \textbf{48.11} & \textbf{17.37} & \textbf{0.605} & \textbf{0.232} & \textbf{0.886} & \textbf{0.041} & \textbf{45.69} \\             
    \midrule
    
    \multirow{3}{*}{4} & FSTH & -- & -- & -- & -- & -- & -- & -- & -- & -- & -- & -- & -- \\
    & LPD & -- & -- & -- & 0.755 & 0.069 & 79.67 & -- & -- & -- & 0.909 & 0.058 & 38.81 \\
    
    & Ours & \textbf{18.90} & \textbf{0.638} & \textbf{0.192} & \textbf{0.909} & \textbf{0.039} & \textbf{43.19} & \textbf{19.33} & \textbf{0.661} & \textbf{0.171} & \textbf{0.930} & \textbf{0.037} & \textbf{34.31}\\
    \midrule

    \multirow{3}{*}{8} & FSTH & 17.86 & 0.600 & 0.225 & 0.836 & 0.046 & 46.38 & 18.35 & 0.647 & 0.218 & 0.899 & 0.044 & 45.15 \\
    & LPD & -- & -- & -- & 0.760 & 0.068 & 77.00 & -- & -- & -- & 0.922 & 0.056 & 35.87 \\
    & Ours & \textbf{19.18} & \textbf{0.645} & \textbf{0.186} & \textbf{0.917} & \textbf{0.039} & \textbf{44.23} & \textbf{19.65} & \textbf{0.675} & \textbf{0.160} & \textbf{0.940} & \textbf{0.036} & \textbf{32.54}\\
    \midrule

    \multirow{3}{*}{32} & FSTH & 18.66 & 0.613 & 0.207 & 0.843 & 0.044 & 44.85 & 19.69 & 0.686 & 0.171 & 0.927 & 0.041 & 33.69 \\
    & LPD & -- & -- & -- & 0.769 & 0.066 & 63.47 & -- & -- & -- & 0.935 & 0.054 & 33.96 \\
    & Ours & \textbf{19.35} & \textbf{0.650} & \textbf{0.182} & \textbf{0.921} & \textbf{0.037} & \textbf{43.32} & \textbf{19.98} & \textbf{0.690} & \textbf{0.146} & \textbf{0.948} & \textbf{0.038} & \textbf{28.26}\\
    \midrule
    
    \end{tabular}
    \vspace{-0.2cm}
    \label{tab:baselines}
\end{table*}
% >>>>>>>>>>>>>>>>>>>>>>>>>>>>>
% <<<<<<<<<<<<<<<<<<<<<<<<<<<<<<<<<<<
\begin{figure*}[t!]
    \centering
    \includegraphics[width=0.92\linewidth]{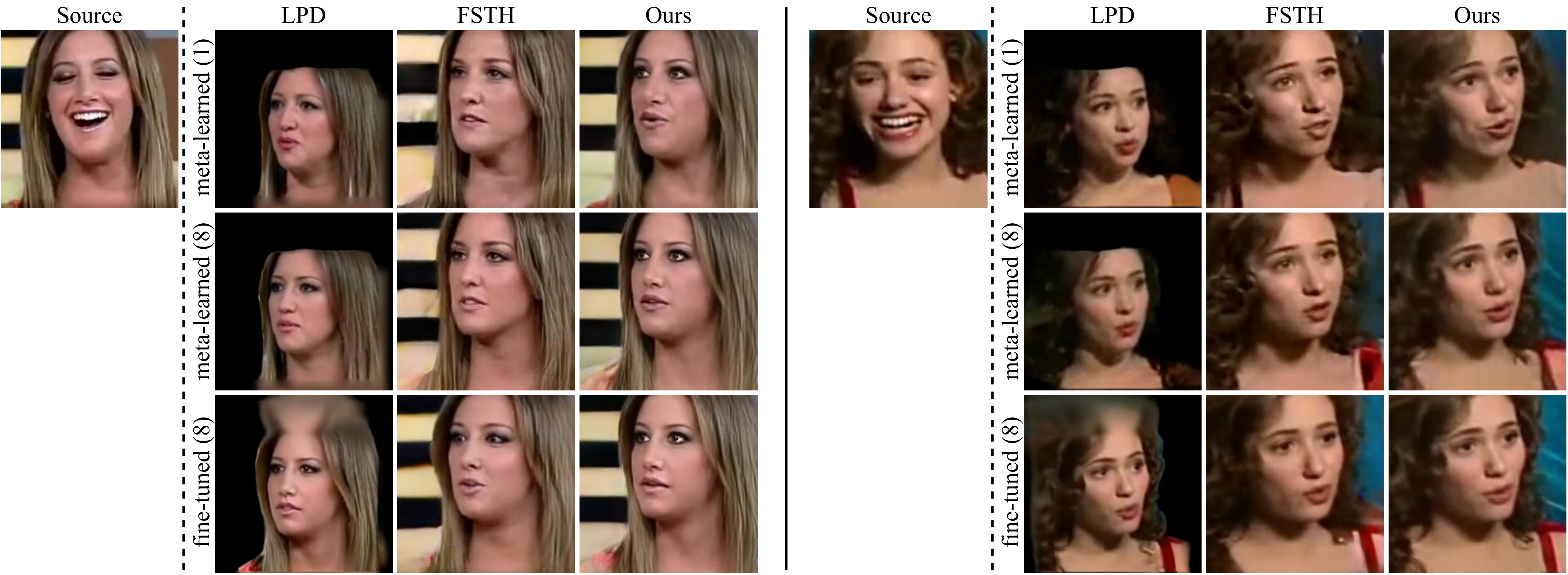}
    \caption{Extending Figure 5 of the main paper. More results comparing our performance to the few-shot baselines with respect to increasing the the K-shot inputs and applying subject fine-tuning.}
    \vspace{-0.2cm}
    \label{fig:var_k_and_ft_more}
\end{figure*}

% >>>>>>>>>>>>>>>>>>>>>>>>>>>>>>>>>>>
\subsubsection{More comparative evaluation}
We report the quantitative details for the effect of increasing the number of K-shot inputs, as well as the effect of subject fine-tuning in Table~\ref{table:var_k_and_ft}.
We observe similar conclusions to those obtained from Figure 6 in the main paper.
LPD~\cite{burkov2020neural} performs very poorly in the meta-learned setting, and only outperforms the FSTH baseline~\cite{zakharov2019few} in the subject fine-tuning setting.
On the other hand, our method consistently outperforms the baselines in all metrics across different settings.
Furthermore, the performance of our method at $K=4$ is on-par with or outperforms the baselines evaluated at $K=32$ across all metrics.
Since the LPD~\cite{burkov2020neural} baseline does not predict the background and re-crops the input/output frames, we subtract the background and compare with their corresponding cropped ground truths for quantitative analysis.
We also exclude LPD from frame reconstruction evaluation since its outputs do not align with the rest of the methods.
Also, the authors of FSTH~\cite{zakharov2019few} only provide their output for $K = \{1,8,32\}$ and they did not release their code. Therefore, we don't report their performance for $K=4$.

% --------------------------------------------------------------
% <<<<<<<<<<<<<<<<<<<<<<<<<<<<<<<
\begin{figure*}[t!]
    \centering
    \includegraphics[width=0.99\linewidth]{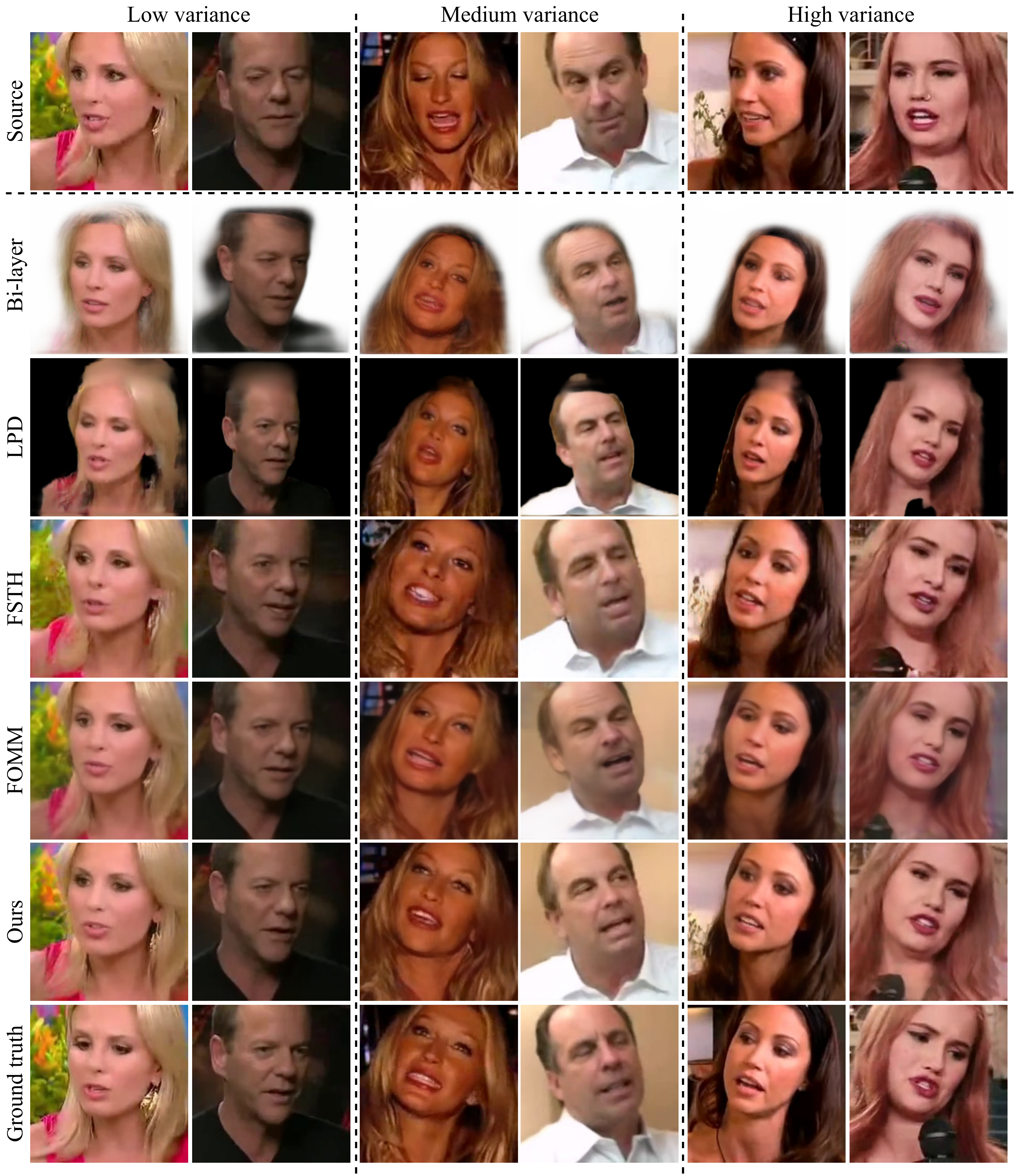}
    \caption{Extending Figure 4 of the main paper, showing more qualitative comparisons in the single-shot setting. We show three sets of examples representing low, medium and high variance between the source and target poses. Our method is more robust to pose variations than the baselines.}
    % \vspace{-0.2cm}
    \label{fig:qual_cmp_more}
\end{figure*}

% >>>>>>>>>>>>>>>>>>>>>>>>>>>>>>>
\subsubsection{More qualitative comparisons}
Here, we expand our qualitative comparisons of the main paper in both the multi-shot and single-shot settings.
Figure~\ref{fig:var_k_and_ft_more} extends Figure 5 of the main paper. It shows more examples comparing the effect of increasing the K-shot inputs and applying subject fine-tuning between our method and the baselines.
Figure~\ref{fig:qual_cmp_more} shows more comparisons in the single-shot setting as Figure 4 in the main paper.
% --------------------------------------------------------------

% <<<<<<<<<<<<<<<<<<<<<<<<<<<<<<<
\begin{figure*}[t!]
    \centering
    \includegraphics[width=0.8\linewidth]{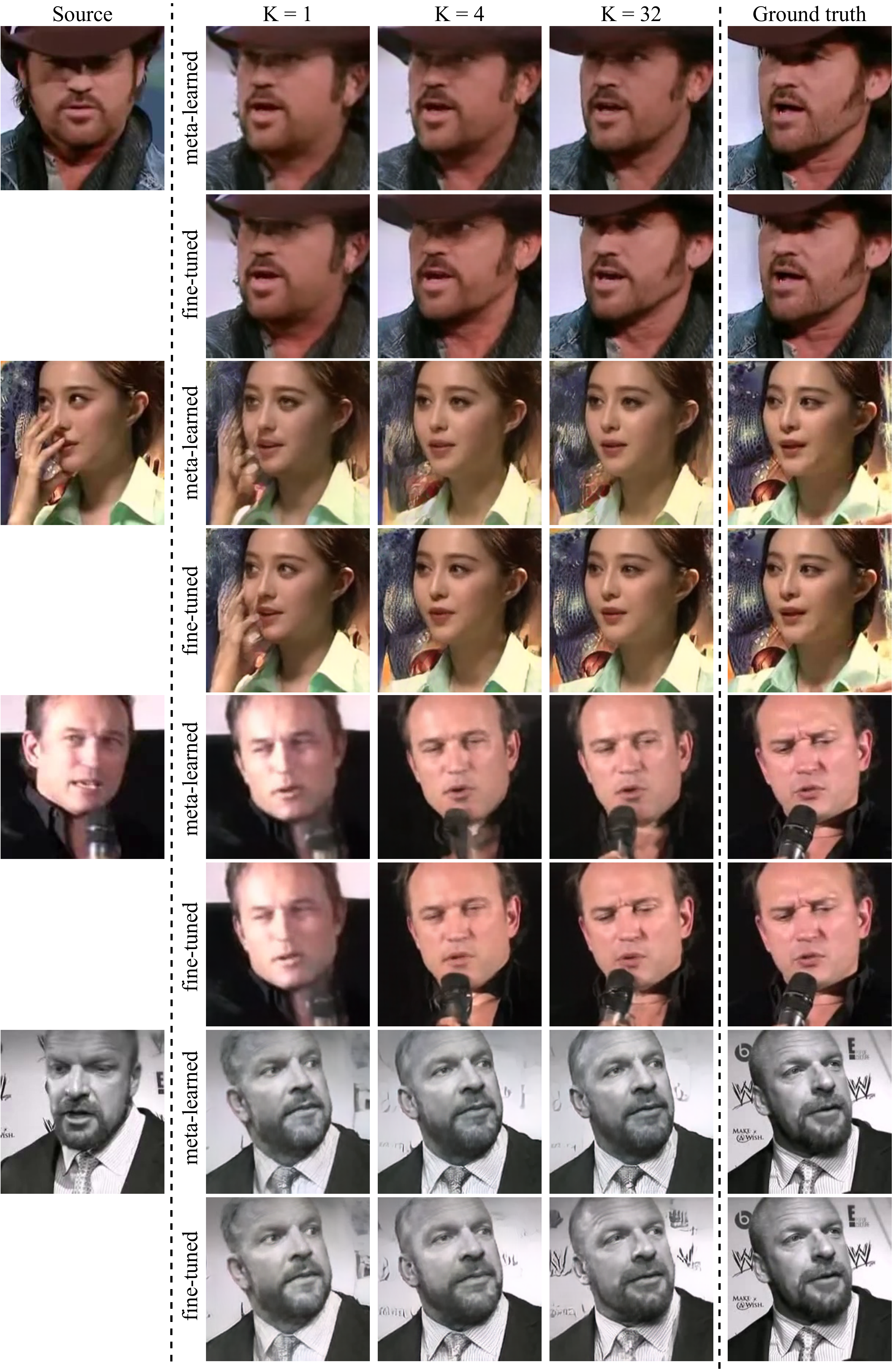}
    \caption{Qualitative results of our method showing the gains of increasing the number of K-shot inputs and applying subject fine-tuning.}
    % \vspace{-0.2cm}
    \label{fig:qual_cmp_ours}
\end{figure*}

% >>>>>>>>>>>>>>>>>>>>>>>>>>>>>>>
% <<<<<<<<<<<<<<<<<<<<<<<<<<<<<<<<<<<<<<<<<<
\begin{figure*}[ht]
    \vspace{-0.3cm}
    \centering
    \includegraphics[width=0.88\linewidth]{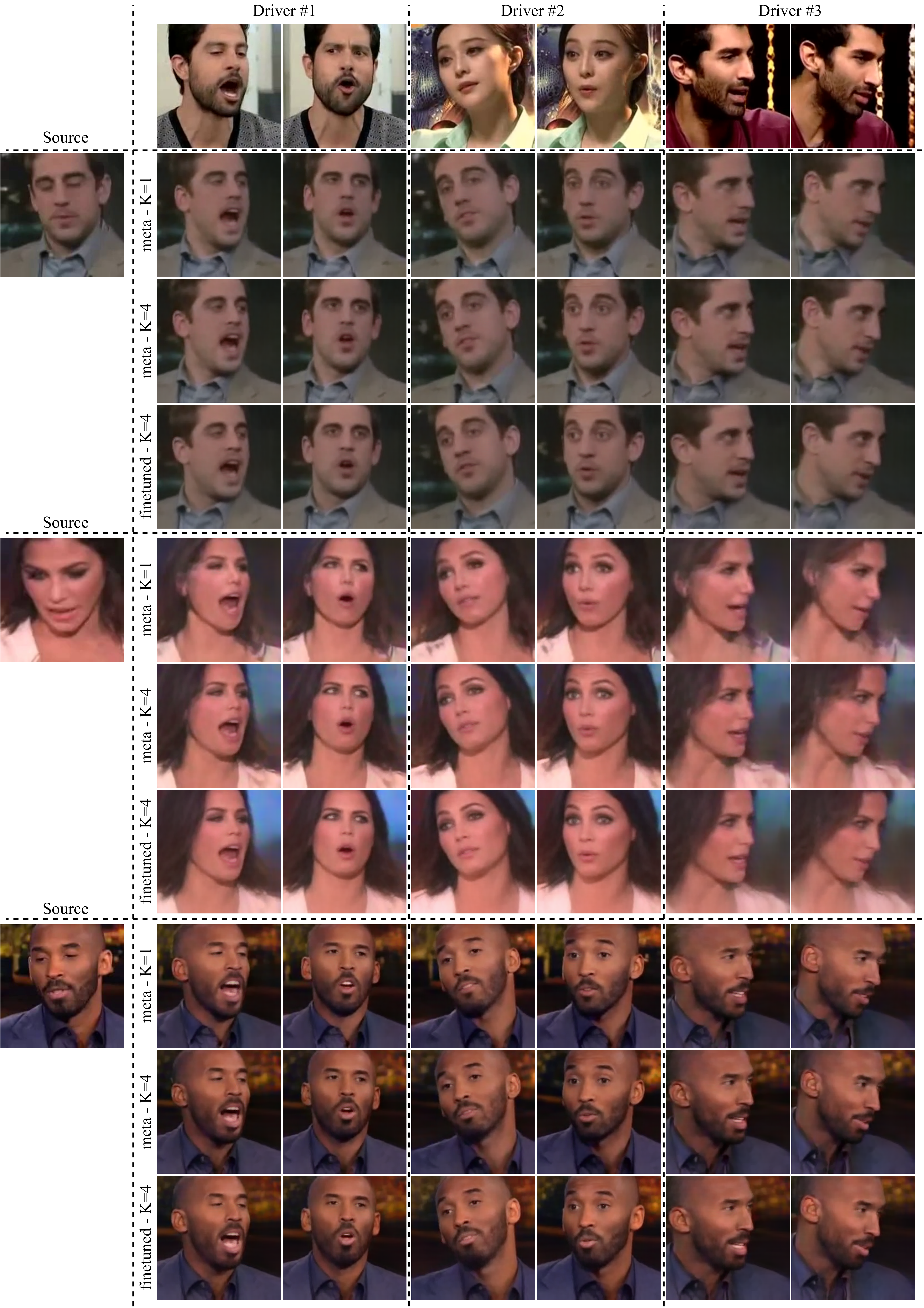}
    \caption{Expanding Figure 7 of the main paper by showing the same reenactment results in the single and 4-shot settings. Our model extrapolates well to challenging poses and expressions even with a single-shot input (shown in source), while preserving the source identity.}
    % \vspace{-0.4cm}
    \label{fig:reenactment_var_k_and_ft}
\end{figure*}

\begin{figure*}[ht]
    \vspace{-0.3cm}
    \centering
    \includegraphics[width=0.95\linewidth]{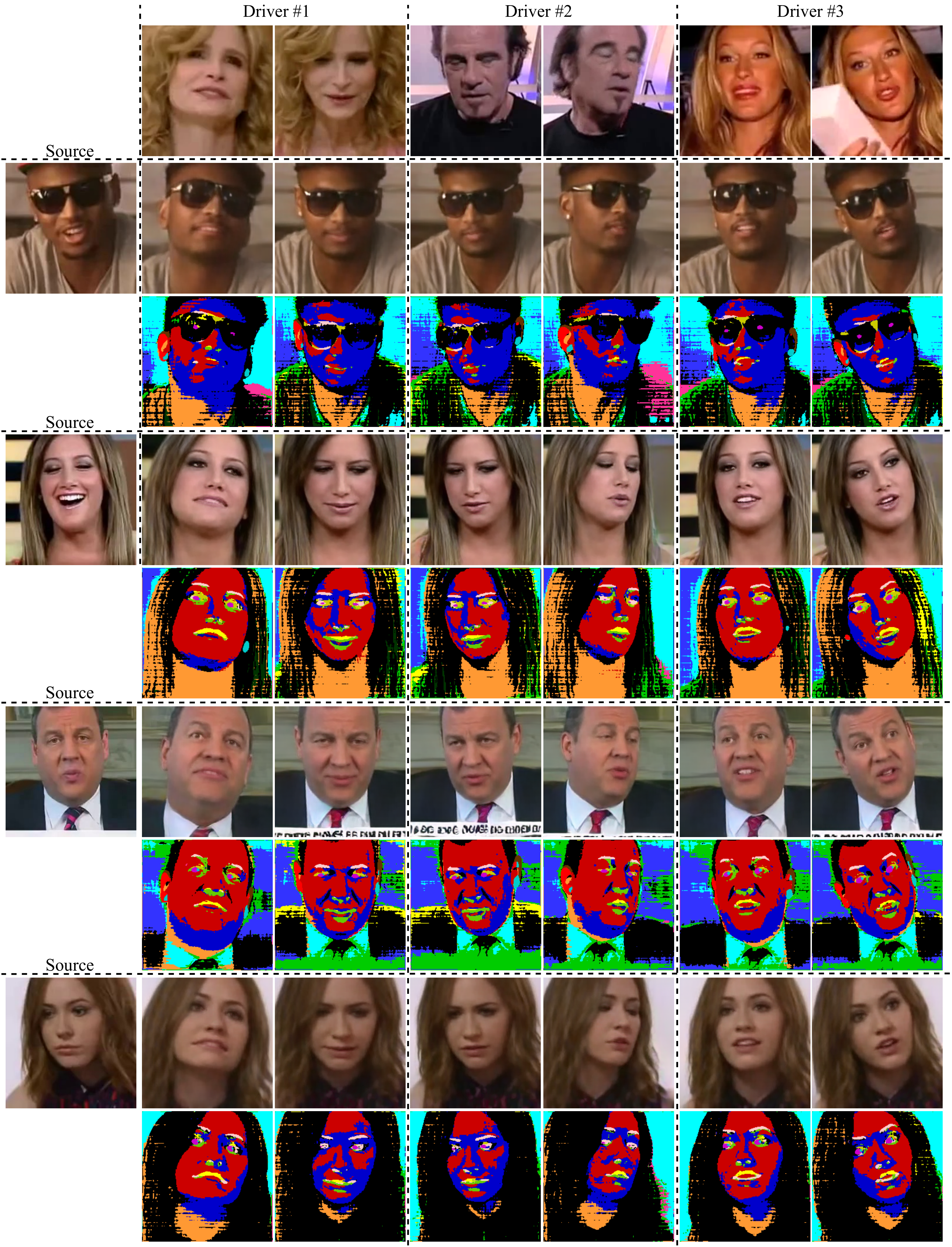}
    \caption{More cross-subject reenactment results with different driving identities. Results are shown for our \emph{meta-learned} model without any fine-tuning, and using 32-shot inputs. We also show the corresponding latent spatial layout maps.}
    \vspace{-0.4cm}
    \label{fig:reenactment_more}
\end{figure*}

% >>>>>>>>>>>>>>>>>>>>>>>>>>>>>>>>>>>>>>>>>>
\subsection{More qualitative results}
\label{sec:var_k_and_ft_qualitative}
We show more qualitative results of our method showing the effect of increasing the K-shot inputs, and the effect of applying the subject fine-tuning in Figure~\ref{fig:qual_cmp_ours}.
We observe that we get a noticeable improvement when we increase K from 1 to 4.
The visual gain from increasing K further starts to saturate, although quantitative metrics generally keep improving (\eg Table~\ref{table:var_k_and_ft}).
While increasing K beyond 4 still leads to better visual results in general, we observe that the most improvement focuses on the background and clothes reconstruction, with slight improvements to sharpness and color matching as well. 
% We observe that increasing K beyond 4 frames mainly improves the background and clothes reconstruction, while also slightly improving sharpness and color matching.
Subject fine-tuning further improves the sharpness and better reconstructs the background details.
% - quantitatively, we get better as we increase K and as we fine-tune
% - qualitatively validate this. 
% - we observe that we get a noticeable improvement when we increase K from 1 to 4.
% - However, as we increase K further, the visual quality improves mainly for background and clothes reconstruction and slight improvements for color matching and sharpness.
% performance starts to saturate visually, although quantitative metrics generally keep improving.
% --------------------------------------------------------------

\begin{figure}[t!]
    \centering
    \includegraphics[width=0.92\linewidth]{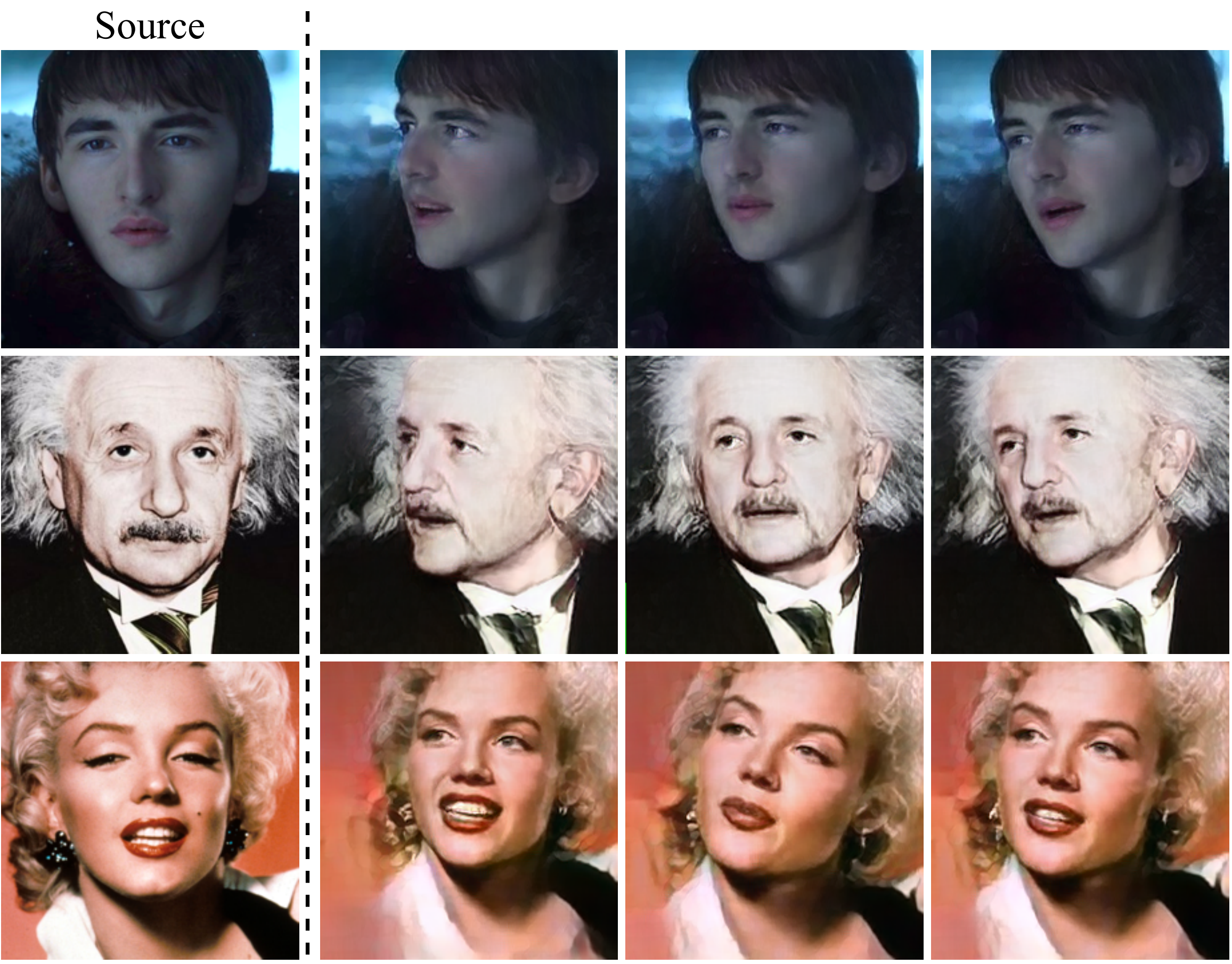}
    \caption{Qualitative results on subjects not belonging to VoxCeleb2.}
    \vspace{-0.2cm}
    \label{fig:wild}
\end{figure}

\subsection{More reenactment results}
\label{sec:reenactment_more}
We first expand Figure 7 of the main paper by showing the same reenactment results but for the single-shot meta-learned setting and the 4-shot inputs in both the meta-learned and subject fine-tuned settings in Figure~\ref{fig:reenactment_var_k_and_ft}.
The results show that even in the single-shot meta-learned setting, our model does a pretty good job extrapolating the input image (source) to challenging poses and expressions, while preserving the source identity.
Increasing the input shots to 4 leads to a noticeable visual improvement, and fine-tuning further leads to slight improvements, most notable in the female source (middle example).
These results show that our method does not require many frames to produce realistic and identity preserving results.
For video comparison with the baselines, please refer to the supplementary video.

We also extend Figure 7 of the main paper by showing more reenactment results in Figure~\ref{fig:reenactment_more}. We also show the predicted spatial layouts corresponding to the outputs.
The predicted spatial layouts may be less interpretable than traditional semantic segmentations, but they seem to encode more information and capture accurate details about the face shape.

Additionally, we perform out-of-domain reenactment using source subjects not present in the VoxCeleb2 dataset. % in the single-shot scenario.
Some qualitative results are shown in Figure~\ref{fig:wild}.
Our approach can synthesize realistic novel views given only a single-shot input, although in some cases it shows a bit of an identity gap.
% It can be seen that even though the subjects are unseen, our approach can synthesize novel views even with a single input frame. Although impressive, it does in some cases not preserve the background properly.
%
% \noindent
% \hl{TODO}\\
% - effect of varying K and fine-tuning.\\
% - Our model does a pretty descent job extrapolating to challenging novel poses and expressions from a single-shot input. 
% - more reenactment results.\\
% - for video comparisons, please refer to the supplementary video.\\
% - monalisa and/or selfies????\\
% --------------------------------------------------------------

% <<<<<<<<<<<<<<<<<<<<<<<<<<<<<
\begin{figure}[t!]
    \centering
    \includegraphics[width=0.92\linewidth]{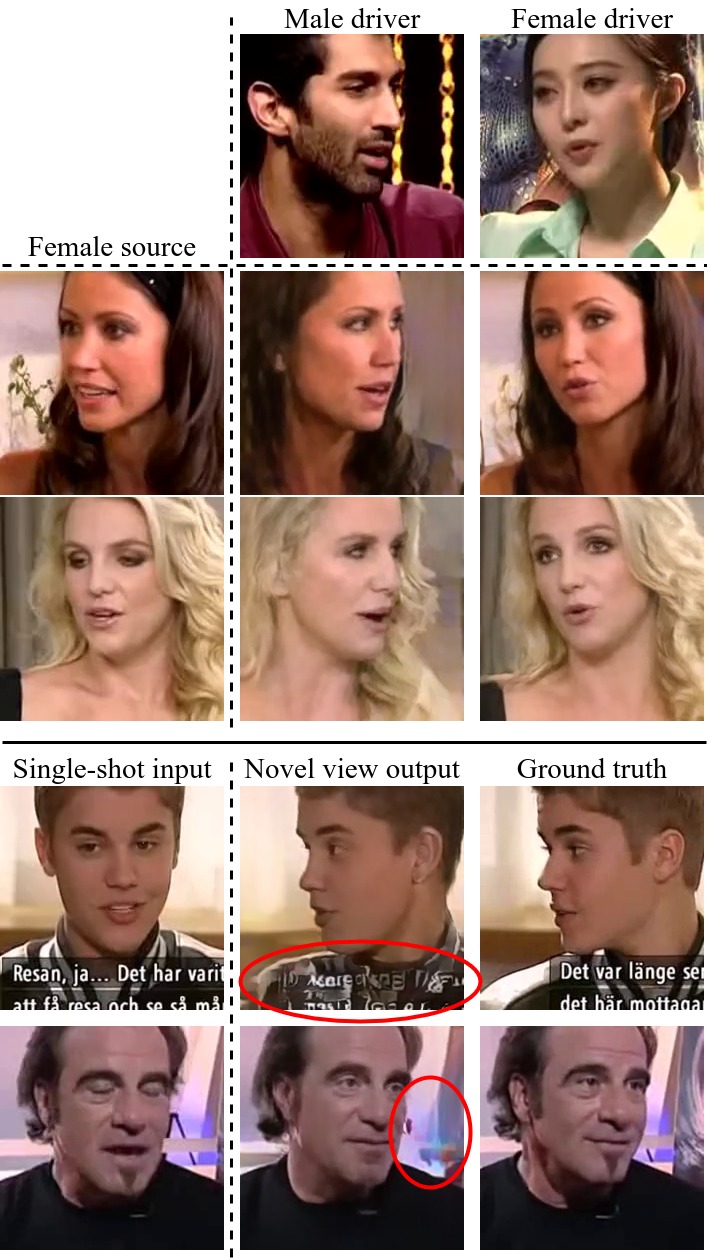}
    \caption{Example limitations. Top: male-to-female reenactment sometimes causes low identity preservation and other visible artifacts. Bottom: our approach cannot faithfully reconstruct the background details.}
    \label{fig:limitations}
\end{figure}

% >>>>>>>>>>>>>>>>>>>>>>>>>>>>>
\subsection{Limitations and failure cases}
\label{sec:limitaitons}
Here we discuss some of the limitations of our approach.
\paragraph{Temporal consistency.}
Similar to previous direct synthesis approaches~\cite{zakharov2019few,burkov2020neural,Zakharov20fast}, our training does not enforce temporal consistency. Therefore the output videos often contain some flickering. Considering the temporal aspect during training (\eg similar to~\cite{wang2018fewshotvid2vid}) could mitigate this problem, but on the expense of higher training cost.
\paragraph{Failure modes for cross-subject reenactment.} We observe that most of the failure cases in cross-subject reenactment are caused by either source subjects with complex backgrounds, or using male drivers to animate female sources (\eg~Figure~\ref{fig:limitations}).
Since complex backgrounds could lead to artifacts in our results, then this signifies that the background information is entangled with the face and identity information.
Learning a better disentangled representation could improve this problem.
On the other hand, the trouble faced with male-to-female reenactment implies that our approach still has some sensitivity to the driver landmarks. While our approach reduces this sensitivity significantly compared to previous baselines, there is still room for improvement.
\paragraph{Background reconstruction}
Direct synthesis approaches, including our method, synthesize the target frame from a compressed latent code.
This compressed bottleneck leads to the loss of some information, especially for the background details.
Figure~\ref{fig:limitations} shows some examples in the single-shot setting. Our method cannot transfer static parts (\eg the closed captions or the background) from the source image to the synthesized view.
Borrowing elements from the warping-based approaches is one direction to better reconstruct static details.
\paragraph{Dataset-induced limitations.} The VoxCeleb2 dataset~\cite{Chung18b} has low resolution videos and is processed to perform zoomed-in center crops that often cut off the top of the head.
Dataset biases are inevitably inherited by the trained models.
Therefore, generating output for out-of-domain inputs requires pre-processing the inputs to have similar properties to the VoxCeleb2 dataset.
% \noindent
% \hl{TODO}\\
% - temporal coherency.\\
% - background details (e.g. gibberish writing).\\
% - Identity preservation for K=1.\\
% - Identity preservation for cross-reenactment: male drivers and female sources.\\
% - sensitivity to voxceleb pre-processing.\\
% --------------------------------------------------------------

\subsection{Ethical concerns}
\label{sec:ethical}
% While we aim in our approach to bridge the gap between artificially synthesized and real sequences, it raises ethical concerns too. A prime example of this is the growing misuse of DeepFakes~\cite{tolosana2020deepfakes, mirsky2021creation}. Additionally, with the increase in ease of access, more and more people can use such models through widely available applications. Current state-of-the-art methods can easily swap identities, expressions as well as face attributes and generate photo-realistic samples. Thus it is important at the same time to have the ability to detect fake content. In this direction recent works like~\cite{marra2018detection, wang2020cnn, zhang2019detecting} have tried to solve the problem of detecting real \vs fake images. Especially interesting is the work by Wang \etal.~\cite{wang2020cnn} which shows that models for fake image detection can be made to generalize well to unseen scenarios. While this is a temporary respite, it is important to continue research in the field of fake image detection to keep on par with the ever improving field of image synthesis as not only do models improve, at the same time ease of use also improves with more people having access to these technologies.
%
While the task of synthesizing realistic \emph{talking heads} has a wide range of applications, it also raises ethical concerns regarding potential misuses of this technology.
A prime example of this is the growing misuse of DeepFakes~\cite{tolosana2020deepfakes, mirsky2021creation}.
Several state-of-the-art methods can easily swap identities, expressions as well as face attributes and generate photo-realistic samples.
Additionally, with the increase in the ease of access to face reenactment models, more and more people can misuse such models through widely available applications.
Thus it is important at the same time to have the ability to detect fake content.
In this direction recent works like~\cite{marra2018detection, wang2020cnn, zhang2019detecting} have tried to solve the problem of detecting real \vs fake images. 
Especially interesting is the work by Wang \etal.~\cite{wang2020cnn} which shows that models for fake image detection can be made to generalize well to unseen scenarios.
While this is a temporary respite, it is important to continue research in the field of fake image detection to keep on par with the ever improving field of image synthesis, as not only do models improve, but also the ease of access to such models grows rapidly.

% --------------------------------------------------------------

\end{document}